\documentclass[usenames,dvipsnames,sigconf]{acmart}

\usepackage[nolist]{acronym}
\AtBeginDocument{%
 \providecommand\BibTeX{{%
    \normalfont B\kern-0.5em{\scshape i\kern-0.25em b}\kern-0.8em\TeX}}}

\usepackage{soul}
\usepackage{color}
\usepackage{wallpaper}
\usepackage{graphicx}
\usepackage{cleveref}
\usepackage{numprint}
\usepackage{paralist}
\usepackage{xspace}
\usepackage[lofdepth,lotdepth]{subfig}

\definecolor{orangeX}{rgb}{1,.5,0}
\definecolor{blueX}{rgb}{.2, .59, .88}
\definecolor{purpleX}{rgb}{.294118, 0, .509804}
\definecolor{greenX}{rgb}{.421, .578, .241}
\definecolor{bole}{rgb}{0.47, 0.27, 0.23}
\definecolor{mypink3}{cmyk}{0, 0.7808, 0.4429, 0.1412}
\definecolor{mygray}{gray}{0.6}
	
\newcommand{\ds}{\textsc{DeepSpectrum\,}}
\newcommand{\wtv}{\textsc{Wav2Vec2.0\,}}
\newcommand{\vit}{\textsc{ViT-FER\,}}
\newcommand{\facenet}{\textsc{FaceNet512\,}}
\newcommand{\opensmile}{\textsc{openSMILE}}

\newcommand{\pyfeat}{\textsc{Py-Feat\,}}

\newcommand{\eg}{e.\,g., }
\newcommand{\ie}{i.\,e., }

\copyrightyear{2024}
\acmYear{2024}
\setcopyright{acmlicensed}\acmConference[MuSe '24]{Proceedings of the 5th Multimodal Sentiment Analysis Challenge and Workshop: Social Signal Quantification and Humor Recognition}{October 28, 2024}{Melbourne, Australia}
\acmBooktitle{Proceedings of the 5th Multimodal Sentiment Analysis Challenge and Workshop: Social Signal Quantification and Humor Recognition (MuSe '24), October 28, 2024, Melbourne, Australia}
\acmPrice{15.00}
\acmDOI{10.1145/xxxxxxx.xxxxxxx}
\acmISBN{xxx-x-xxxx-xxxx-x/xx/xx}

\begin{document}
\title[MuSe 2024: Baseline Paper]{
The MuSe 2024 Multimodal Sentiment Analysis Challenge: \\Social Perception and Humor Recognition}

\author{Shahin Amiriparian}
\affiliation{%
  \institution{Chair of Health Informatics\\Klinikum rechts der Isar\\ Technical University of Munich}
  \city{Munich, Germany}}

\author{Lukas Christ}
\affiliation{%
  \institution{Chair of Embedded Intelligence for Healthcare and Wellbeing\\University of Augsburg}
  \city{Augsburg, Germany}}

\author{Alexander Kathan}
\affiliation{%
  \institution{Chair of Embedded Intelligence for Healthcare and Wellbeing\\University of Augsburg}
  \city{Augsburg, Germany}}

\author{Maurice Gerczuk}
\affiliation{%
  \institution{Chair of Embedded Intelligence for Healthcare and Wellbeing\\University of Augsburg}
  \city{Augsburg, Germany}}
  
  \author{Niklas M\"uller}
\affiliation{%
  \institution{Chair of Strategic Management, Innovation, and Entrepreneurship\\University of Passau}
  \city{Passau, Germany}}
  
  \author{Steffen Klug}
\affiliation{%
  \institution{Chair of Strategic Management, Innovation, and Entrepreneurship\\University of Passau}
  \city{Passau, Germany}}

\author{Lukas Stappen}
\affiliation{%
  \institution{Recoro}
  \city{Munich, Germany}}
  
\author{Andreas K\"onig}
\affiliation{%
  \institution{Chair of Strategic Management, Innovation, and Entrepreneurship\\University of Passau}
  \city{Passau, Germany}}

\author{Erik Cambria}
\affiliation{%
  \institution{College of Computing \& Data Science\\
  Nanyang Technological University}
  \city{Singapore, Singapore}}

\author{Bj\"orn W. Schuller}
\affiliation{%
  \institution{Group on Language, Audio, \& Music\\
  Imperial College London}
  \city{London, United Kingdom}}

\author{Simone Eulitz}
\affiliation{%
  \institution{Institute of Strategic Management\\LMU Munich}
  \city{Munich, Germany}}
\renewcommand{\shortauthors}{Shahin Amiriparian et al.}

\settopmatter{printacmref=true}
\copyrightyear{2024}
\acmYear{2024}
\setcopyright{acmlicensed}

\begin{abstract}

The \ac{MuSe} 2024 addresses two contemporary multimodal affect and sentiment analysis problems: 
In the \ac{MuSe-Perception}, participants will predict $16$ different social attributes of individuals such as \textit{assertiveness}, \textit{dominance}, \textit{likability}, and \textit{sincerity} based on the provided audio-visual data. The \ac{MuSe-Humor} dataset expands upon the \ac{Passau-SFCH} dataset, focusing on the detection of spontaneous humor in a cross-lingual and cross-cultural setting. The main objective of \ac{MuSe} 2024 is to unite a broad audience from various research domains, including multimodal sentiment analysis, audio-visual affective computing, continuous signal processing, and natural language processing.
By fostering collaboration and exchange among experts in these fields, the \ac{MuSe} 2024 endeavors to advance the understanding and application of sentiment analysis and affective computing across multiple modalities. This baseline paper provides details on each sub-challenge and its corresponding dataset, extracted features from each data modality, and discusses challenge baselines. For our baseline system, we make use of a range of Transformers and expert-designed features and train \ac{GRU}-\ac{RNN} models on them, resulting in a competitive baseline system. On the unseen test datasets of the respective sub-challenges, it achieves a mean Pearson's Correlation Coefficient ($\rho$) of \textbf{0.3573} for \ac{MuSe-Perception} and an \ac{AUC} value of \textbf{0.8682} for \ac{MuSe-Humor}.
\end{abstract}



\maketitle
\section{Introduction}
In its 5th edition, the \acf{MuSe} proposes two tasks, namely audio-visual analysis of perceived characteristics of individuals and 
cross-cultural humor detection. Each respective sub-challenge employs a distinct dataset.

In the \textbf{first sub-challenge, \ac{MuSe-Perception}}, participants are tasked with training their machine learning models for recognition of perceived characteristics of individuals from video interviews. Audio-visual \textbf{social perception} analysis explores social traits that are important for how we are being perceived by other people~\cite{abele2021navigating}. The perception others have of us can have a significant impact on our relationships with them as well as on our own professional future~\cite{ambady2008first}. At the same time, social perception is a complex phenomenon for which different theories have been brought forward. The Dual Perspective Model (DPM)~\cite{abele_chapter_2014} is based on the dimensions of agency (\ie traits related to goal-achievement such as competence or dominance) and communality (\ie traits referring to social relations such as friendliness or warmth)~\cite{bakan_duality_1966}. The former is stereotypically associated with masculine gender roles, while the latter is traditionally attributed to femininity~\cite{eagly2002role}. 
In this context, the work by~\citet{eulitz_beyond_2021} underscores the significance of understanding how perceived characteristics, particularly agentic traits, influence professional success, including the performance of CEOs in the stock market. Such insights are invaluable for organizations aiming to thrive in dynamic and competitive environments. However, extracting these insights requires advanced tools capable of capturing and analyzing nuanced social signals. Traditional methods of assessing social perception often rely on structured questionnaires or observational techniques to gauge individuals' interpretations and responses to social cues and interactions. In contrast, audio-visual machine learning systems offer scalability, enabling researchers to analyze large datasets efficiently. Moreover, these systems can uncover patterns and correlations that may not be immediately apparent to human observers, leading to deeper insights into the factors influencing social perception. This is where audio-visual machine learning systems play a pivotal role and can offer a holistic approach to understanding social perception by leveraging both auditory and visual cues. They can capture subtle nuances in facial expressions~\cite{kachur2020assessing,sonlu2021conversational,calder2011personality}, body language~\cite{breil202113}, tone of voice~\cite{sonlu2021conversational}, and verbal content~\cite{cutler2022deep,pradhan2020analysis}, providing a rich set of features to analyze. In the context of the \acl{MuSe-Perception}, the \ac{LMU-ELP} dataset (cf.~\Cref{ssec:perception}) presents a unique opportunity to explore the intersection of audio-visual data and social perception.

In the \textbf{second task}, \textbf{\acf{MuSe-Humor}}, participants will train their models to detect humor within German football press conference recordings.

Humor is a ubiquitous yet puzzling phenomenon in human communication. While it is strongly connected with an intention to amuse one's audience~\cite{gkorezis2014leader}, humor has been shown to potentially elicit a wide range of both positive and negative effects~\cite{cann2014assessing}.  Throughout the last three decades, researchers in Affective Computing and related fields have addressed problems pertaining to computational humor, in particular, automatic humor detection~\cite{taylor2004computationally, yang2015humor, pramanick2022multimodal}. Spurred by the advancements in multimodal sentiment analysis, considerable attention has been paid to multimodal approaches to humor detection recently~\cite{bertero2016deep, hasan2021humor}. Such multimodal methods are particularly promising, as humor in personal interactions is an inherently multimodal phenomenon, expressed not just by means of what is said, but also via, \eg gestures and facial expressions.
A plethora of datasets dedicated to multimodal humor recognition exists~\cite{hasan2019ur, mittal2021so, wu2021mumor}. However, they are typically built from recordings of staged contexts, \eg TV shows or TED talks, thus potentially missing out on spontaneous, \textit{in-the-wild} aspects of humorous communication. Furthermore, several such datasets utilize audience laughter as a proxy label for humor, thus taking the risk of reducing the complex concept of humorous communication to the mere delivery of (scripted) punchlines. The \ac{Passau-SFCH} dataset~\cite{christ2022multimodal} seeks to mitigate these issues by providing videos from a semi-staged context, namely press conferences, that have been labeled manually for humor.

\ac{MuSe-Humor}  adds another layer of complexity to the humor recognition task by introducing a cross-cultural scenario. More specifically, the training data consists of German recordings, while the test data is composed of English videos. Whereas empirical studies have been conducted on cross-cultural similarities and differences in humor~\cite{priego2018smiling, ladilova2022humor}, this problem has not received much attention from the machine learning domain. To the best of our knowledge, last year's edition of \ac{MuSe-Humor} was the first to introduce this task to the community, giving rise to a range of different systems proposed by the challenge's participants~\cite{grosz2023discovering, xu2023humor, li2023jtma, yu2023mmt, xie2023multimodal}.

This year's edition of \ac{MuSe-Humor} maintains the same data and data partitions as last year's~\cite{christ2023muse}. The training set comprises recordings from German football press conferences, while the unseen test set exclusively features press conferences conducted in English, thus setting the stage for a cross-cultural, cross-lingual evaluation scenario. 
For the training partition, we utilize the German \acf{Passau-SFCH} (\textbf{\ac{Passau-SFCH}}) dataset~\cite{christ2022multimodal}, previously featured in the 2022 and 2023 editions of MuSe~\cite{xu2022hybrid, chen2022integrating, li2022hybrid, grosz2023discovering, xie2023multimodal, yu2023mmt}. For the unseen test partition, we expanded the \ac{Passau-SFCH} dataset by incorporating press conference recordings delivered by seven distinct coaches from the English Premier League spanning from September 2016 to September 2020. Both the training and test partitions exclusively contain recordings where the respective coach is speaking, contributing to an overall duration exceeding 17 hours. Although the original videos are labeled according to the \ac{HSQ} framework introduced by Martin et al.~\cite{martin2003individual}, the focus in this year's \ac{MuSe-Humor} challenge is on binary prediction, specifically detecting the presence or absence of humor.

\begin{table}[t!]
\renewcommand{\arraystretch}{1.2}
\footnotesize
  \caption{Statistics for each sub-challenge dataset. Included are the number of unique subjects (\#), and the video durations formatted as h:mm:ss, and the total number of subjects along with the overall duration of all recordings in each dataset.\label{tab:partitioning}}
 \resizebox{\linewidth}{!}{
  \begin{tabular}{lrcrc}
    \toprule
      & \multicolumn{2}{c}{\textbf{\ac{MuSe-Perception}}} & \multicolumn{2}{c}{\textbf{\ac{MuSe-Humor}}}\\
     \cmidrule(lr){2-3} \cmidrule(lr){4-5}
    Partition & \# & Duration & \# &  Duration \\
    \midrule
    Train   & 59 & 0\,:30\,:13 & 7 & 7\,:44\,:49 \\
    Development  &  58 & 0\,:29\,:12 & 3 & 3\,:06\,:48 \\
    Test    &  60 & 0\,:30\,:10 & 6 & 6\,:35\,:16 \\
    \midrule
    $\sum$    & 177 & 1\,:29\,:35 & 16 & 17\,:26\,:53 \\
  \bottomrule
\end{tabular}
}
\end{table}

The sub-challenges presented in \ac{MuSe} 2024 are designed to captivate a broad audience, drawing the interest of researchers spanning numerous domains, including multimodal sentiment analysis, affective computing, natural language processing, and signal processing. \ac{MuSe} 2024 provides participants with an ideal platform to apply their expertise by leveraging multimodal self-supervised and representation learning techniques, as well as harnessing the power of generative AI and \acp{LLM}. 

Each sub-challenge is characterized by its unique datasets and prediction objectives, providing participants with an opportunity to delve into specific problem domains while simultaneously contributing to broader research endeavors. By serving as a central hub for the comparison of methodologies across tasks, \ac{MuSe} facilitates the discovery of novel insights into the effectiveness of various approaches, modalities, and features within the field of affective computing.

In~\Cref{sec:challenges}, we provide a comprehensive overview of the outlined sub-challenges, detailing their corresponding datasets and the challenge protocol. Following this,~\Cref{sec:features} elaborates on our pre-processing and feature extraction pipeline, along with the experimental setup utilized to compute baseline results for each sub-challenge. Subsequently, the results are showcased in~\Cref{sec:results}, leading to the conclusion of the paper in~\Cref{sec:conclusion}.



\section{The Two Sub-Challenges}\label{sec:challenges}
In this section, we describe each sub-challenge and dataset in detail and provide a description of the challenge protocol. 

\subsection{The \acl{MuSe-Perception}}
\label{ssec:perception}

For the \textbf{first task}, \textbf{Social Perception Sub-Challenge \textsc{(MuSe-Perception)}}, we introduce the novel \ac{LMU-ELP} dataset, which consists of audio-visual recordings  
of US executives, specifically the highest-ranking executives of listed firms - chief executive officers (CEOs). In our dataset, these CEOs present their firms to potential investors before taking their firms public, \ie selling shares in the stock market for the first time. The goal of this challenge is to predict the agency and communiality of each individual on a $16$-dimensional Likert scale ranging from $1$ to $7$. The target labels have been selected based on an established measure in social psychology (\acl{BSRI} scale~\cite{bem1981manual}) to assess perceptions of agency and commonality, the basic dimension of the Dual Perspective Model ~\cite{abele_chapter_2014}, including \emph{aggressiveness, arrogance, assertiveness, confidence, dominance, independence, leadership qualities,} and \emph{risk-taking propensity} (pertaining to agency), as well as attributes like \emph{collaboration, enthusiasm, friendliness, good-naturedness, kindness, likability, sincerity,} and \emph{warmth} (associated with communality). Using Amazon Mechanical Turk (MTurk), $4304$ annotators have labeled all dimensions of our data, comprising $177$ CEOs. The dataset stands out for its comprehensive coverage of $16$ distinguished labels of each individual, offering new perspectives in multimodal sensing of social signals. To the best of our knowledge, \ac{LMU-ELP} is the first multimodal dataset that provides detailed insights into the nuanced dimensions of gender-associated attributes. 
\ac{MuSe-Perception} participants are encouraged to explore multimodal machine learning methods to automatically recognize CEOs' perceived  characteristics. Each team will submit their $16$ predictions. The evaluation metric is Pearson’s correlation coefficient ($\rho$), and the mean of all $16$ $\rho$ values is set as the challenge’s evaluation criterion.

\subsection{The Cross-Cultural Humor Sub-Challenge\label{ssec:humor}} 

The training partition provided in \ac{MuSe-Humor} comprises videos of $10$ football coaches from the German Bundesliga, all of them native German speakers. The test set, in contrast, features $6$ English Premier League coaches from $6$ different countries (Argentine, England, France, Germany, Portugal, Spain), only one of them being a native speaker. All subjects are male and aged between $30$ and $53$ years (training partition) and $47$ to $57$ years (test), respectively. $7$ of the $10$ German coaches are utilized as the training set, whereas the remaining $3$ are used as the development set.~\Cref{tab:partitioning} provides further statistics of the dataset.

The videos are cut to only include segments in which the coaches are speaking. In addition to the recordings, manual transcriptions with corresponding timestamps are provided. 

The videos in \ac{Passau-SFCH} are originally labeled according to the HSQ, a two-dimensional model of humor proposed by~\citet{martin2003individual}. From these annotations, the gold standard is created as elaborated in~\cite{christ2022multimodal}, such that each data point corresponds to a $2$\,s frame with a binary label. Overall, humorous segments make up $4.38$\,\% of the training segments, $2.81$\,\% of the development segments, and $6.17$\,\% of the segments used for the test. 

Same as in previous editions of \ac{MuSe-Humor}, \ac{AUC} serves as the evaluation metric.

\subsection{Challenge Protocol}
To join the challenge, participants must be affiliated with an academic institution and complete the EULA found on the \ac{MuSe} 2024 homepage\footnote{\href{https://www.muse-challenge.org/challenge/participate}{https://www.muse-challenge.org/challenge/participate}} homepage. The organizers will not compete in any sub-challenge. During the contest, participants upload their predictions for test labels on the CodaBench platform. Each sub-challenge allows up to $5$ predictions, where the best among them determines the rank on the leaderboard. All teams are encouraged to submit a paper detailing their experiments and results. Winning a sub-challenge requires an accepted paper. All papers are subject to a double-blind peer-review process.

\section{Baseline Approaches}\label{sec:features}
To support participants in time-efficient model development and reproducing challenge baselines, we provide a set of expert-designed (explainable) and Transformers'-based features. For the \ac{MuSe-Perception} task, we provide 6 feature sets, including three vision-based and three audio-based features. Additionally, for the \ac{MuSe-Humor} sub-challenge, we provide an extra feature set tailored for the text modality, totaling 7 feature sets.

\subsection{Pre-processing}
We split each dataset into three partitions for training, development, and testing, ensuring that the distribution of target labels and the overall length of recordings are balanced across each partition. Moreover, we maintained speaker independence throughout all partitions.

For the \ac{LMU-ELP} recordings (each CEO recording has a fixed length of $30$ seconds), we have removed data points in which the faces of multiple persons are presented, ensuring that each recording solely contains the CEO for which the labeling has been conducted. Furthermore, we have eliminated video recordings with fewer than 5 face frames available. For the audio modality of \ac{LMU-ELP} dataset, we removed the background noise using the free online tool \textit{Vocal Remover}\footnote{\hyperlink{https://vocalremover.org/}{https://vocalremover.org/}}.

From the test partition of the \ac{Passau-SFCH} dataset, we have manually removed video clips if the coaches did not speak English. Additionally, we discarded clips with poor audio quality.

\subsection{Audio}
Prior to extracting audio features, we normalize all audio files to $-3$\,dB and convert them to mono format, with a sampling rate of $16$\,kHz and a bit depth of $16$. Subsequently, we employ the \opensmile{} toolkit~\cite{eyben2010opensmile} to compute handcrafted features. Additionally, we generate high-dimensional audio representations using both \ds{}~\cite{Amiriparian17-SSC} and a modified version of \wtv{}~\cite{baevski2020wav2vec}. 
Both systems have demonstrated their efficacy in audio-based \ac{SER} and sentiment analysis tasks~\cite{Amiriparian22-DAP,Gerczuk22-EAT,Schuller21-TI2,Amiriparian17-SAU}.

\subsubsection{\acs{eGeMAPS}}
\label{ssec:egemaps}

We utilize the \opensmile{} toolkit~\cite{eyben2010opensmile} to extract $88$ dimensional \ac{eGeMAPS} features~\cite{eyben2015geneva}, which have shown to be robust for sentiment analysis and \ac{SER} tasks~\cite{baird2019can,vlasenko2021fusion,xu2022hybrid,park2022towards}. We employ the standard configuration for each sub-challenge and extract features using a window size of $2000$\,ms and a hop size of $500$\,ms.

\subsubsection{\ds}
Utilizing \ds{} \cite{Amiriparian17-SSC}, we harness (more conventional) deep \ac{CNN}-based representations from audio data. Our approach involves the initial generation of Mel-spectrograms for each audio file, employing a window size of $1000$\,ms and a hop size of $500$\,ms, with a configuration of $128$ Mels and utilizing the \emph{viridis} color mapping. These spectrogram representations are then fed into \textsc{DenseNet121}. We extract a $1024$-dimensional feature vector from the output of the last pooling layer. The effectiveness of \ds{} has been validated across various speech and audio recognition tasks \cite{park2022towards,park2023muse,Ottl20-GSE, Amiriparian17-SAU,Amiriparian20-TCP}.

\subsubsection{\wtv}
In recent years, there has been a major interest in self-supervised pretrained Transformer models in computer audition~\cite{ssl-survey}. An exemplary foundation model is \wtv{}~\cite{baevski2020wav2vec}, which has found widespread application in~\ac{SER} tasks~\cite{pepino21_interspeech, morais2022speech}. Given that all sub-challenges are affect-related, we opt for a large version of \wtv{} fine-tuned on the MSP-Podcast dataset specifically for emotion recognition~\cite{wagner2022dawn}\footnote{\href{https://huggingface.co/audeering/wav2vec2-large-robust-12-ft-emotion-msp-dim}{https://huggingface.co/audeering/wav2vec2-large-robust-12-ft-emotion-msp-dim}}. We extract deep features from an audio signal by averaging its representations in the final layer of this model, yielding $1024$-dimensional embeddings.

For \ac{MuSe-Humor}, we extract $2$ Hz features by sliding a $3000$\,ms window over each audio file, with a step size of $500$\,ms, and for \ac{MuSe-Perception}, a $2000$\,ms window size (with a hop size of $500$\,ms) is applied. In the previous edition of \ac{MuSe}, participants frequently employed \wtv{} features, demonstrating their efficacy in capturing nuanced audio characteristics for the affective computing tasks~\cite{li2023temporal,grosz2023discovering,yu2023mmt,xie2023multimodal,li2023jtma}.

\subsection{Video}
To compute the visual modality baseline, we solely extract representations from subjects' faces. In order to do so, we utilize \ac{MTCNN} to isolate faces from video recordings in both datasets. Subsequently, we obtain \acp{FAU}, \facenet, and \vit representations from each face image.

These features already proved to be useful in affect and sentiment analysis tasks. For example, \citet{li2023jtma} used \ac{FAU}, \facenet, and ViT features, among others, and integrated them into a multimodal feature encoder module to perform binary humor recognition.
Similarly, recent works (\eg \cite{xie2023multimodal,park2023muse}) conducted experiments applying these features, achieving an overall best result when fusing them with other features from the audio and text modality. 
Another study by~\citet{yi2023exploring} comes to similar conclusions, presenting a multimodal transformer fusion approach in which they used \ac{FAU} and ViT features.

\subsubsection{\acs{MTCNN}}
We utilize the \ac{MTCNN} face detection model~\cite{zhang2016mtcnn} to extract images of the subjects' faces.

In numerous videos from both datasets, multiple individuals are often visible, yet only the CEOs in the \ac{MuSe-Perception} dataset or the coaches in the \ac{MuSe-Humor} dataset are relevant. Initially, we aim to automatically filter out the CEO's or coach's face in each video through clustering of face embeddings. Subsequently, we manually refine the obtained face sets, retaining only the one corresponding to the CEO or coach. 
After the extraction of face images, we employ \pyfeat, \facenet, and \vit for feature extraction.

\subsubsection{Facial Action Units (FAU)}
\acp{FAU}~\cite{ekman1978facial} offer a transparent method for encoding facial expressions by linking them to the activation of specific facial muscles. Given the rich emotional information conveyed by facial expressions, \acp{FAU} have received substantial interest in the sentiment analysis and affective computing community~\cite{zhi2020comprehensive}. We employ the pyfeat library\footnote{\url{https://py-feat.org}} to automatically estimate the activation levels of $20$ distinct \acp{FAU}.

\subsubsection{\facenet}
We employ the \facenet model~\cite{schroff2015facenet}, which is specially trained for face recognition. Specifically, we utilize the implementation provided in the deepface library~\cite{serengil2020lightface}, yielding a 512-dimensional embedding for each face image.

\subsubsection{Vision Transformer (\vit)}

Additionally, we leverage a finetuned variant of ViT~\cite{dosovitskiy2021an} provided by~\cite{chaudhari2022vitfer}\footnote{\url{https://huggingface.co/trpakov/vit-face-expression}}, adapted to emotion recognition on the FER2013 dataset~\cite{goodfellow2013challenges}. We choose this approach since, contrary to the original ViT model, it is explicitly trained on pictures of faces rather than a wide variety of domains. The last hidden state of the special \verb|[CLS]| token is utilized as the representation of a face image. 

\subsection{Text: Transformers}
This modality is exclusive to the \ac{MuSe-Humor} sub-challenge. Given that \ac{MuSe-Humor} involves training and development data in German but a test set in English, we utilize the multilingual version of BERT~\cite{devlin-etal-2019-bert}\footnote{\href{https://huggingface.co/bert-base-multilingual-cased}{https://huggingface.co/bert-base-multilingual-cased}}, pretrained on Wikipedia entries across $104$ languages, including German and English.  This model has demonstrated effective generalization across different languages~\cite{pires-etal-2019-multilingual}. Specifically, we compute sentence representations by extracting the encoding of the \verb|CLS| token from the model's final layer ($768$-dimensional), which represents the entire input sentence. 

While \textsc{BERT} embeddings have demonstrated to be suitable for the task of humor recognition~\cite{xu2023humor, li2023jtma, grosz2023discovering}, they can only be regarded as a simple baseline. The advent of larger \acp{LLM} such as LLaMa~\cite{touvron2023llama} or Mistral~\cite{jiang2023mistral} opens up new avenues for humor recognition based on the textual modality. We would like to strongly encourage participants to explore the aptitude of (multilingual) \acp{LLM} for \ac{MuSe-Humor}, as the text modality proves to be promising for the problem at hand (cf.~\Cref{ssec:results_humor}). 

\subsection{Alignment}

The \ac{Passau-SFCH} dataset comprises audio, video, and transcripts. To facilitate the development of multimodal models leveraging these features, we align the various modalities with each other and with the labeling scheme specific to the task.

Both audio and face-based features are extracted at a rate of $2$\,Hz, using sliding windows with a step size of $500$\,ms for audio, and sampling faces at a $2$\,Hz rate for the video modality. Deriving sentence-wise timestamps from manual transcripts is done in three steps: firstly, the Montreal Forced Aligner (MFA)~\cite{mcauliffe2017montreal} toolkit is utilized to generate word-level timestamps. Next, punctuation is automatically added to the transcripts using the \verb|deepmultilingualpunctuation| tool~\cite{guhr-EtAl:2021:fullstop}. Finally, the transcripts are segmented into sentences using PySBD~\cite{sadvilkar-neumann-2020-pysbd}, allowing for the inference of sentence-wise timestamps from the word-level timestamps. Subsequently, $2$\,Hz textual features are computed by averaging the embeddings of sentences overlapping with the respective $500$\,ms windows.

As the labels in \ac{Passau-SFCH} correspond to windows of size $2$\,s, the alignment of features for \ac{MuSe-Humor} with annotations isn't direct but can be facilitated.

In \ac{MuSe-Perception}, the labels pertain to entire videos, eliminating the need for alignment with the labels.

\subsection{Baseline Training\label{sec:model}}
We provide participants with a simple baseline system based on a \ac{GRU}-\ac{RNN}. Our model encodes sequential data via a stack of \ac{GRU} layers. The final hidden representation of the last of these layers is taken as the embedding of the entire sequence and fed into two feed-forward layers for classification. For both tasks, hyperparameters, including the number of \ac{GRU} layers, learning rate, and \ac{GRU} representation size, are optimized. To obtain a set of unimodal baselines, the model is optimized and trained for each feature as described in~\Cref{sec:features}. For a simple multimodal baseline, we employ a weighted late fusion approach, averaging the best-performing audio-based model's predictions with those of the best-performing video-based model. The predictions are weighted by the performance of the respective model on the development set. 
Simulating challenge conditions, we conduct all experiments with $5$ fixed seeds for both sub-challenges. We provide the baseline code, checkpoints, and hyperparameter configurations in the accompanying GitHub repository\footnote{\href{https://github.com/amirip/MuSe-2024}{https://github.com/amirip/MuSe-2024}}.

\subsubsection{\ac{MuSe-Perception}}
For each of the $16$ prediction targets in \ac{MuSe-Perception}, a separate model is trained. We observe this approach to be more promising than a multi-label prediction setup, \ie one model predicting all $16$ labels. Consequently, the late fusion approach also fuses the best video and audio models per target label. Nonetheless, we would like to encourage participants to explore combinations of prediction targets. In order to keep the number of experiments in a reasonable range, we optimize the hyperparameters on one target only and employ the configuration thus found for all $16$ labels. As \ac{MuSe-Perception} is a regression task, we choose \ac{MSE} for the loss function.

\subsubsection{\ac{MuSe-Humor}}
In the \ac{Passau-SFCH} dataset, each label pertains to a $2$\,s frame. As the features are extracted in $500$\,ms intervals, data points in the \ac{MuSe-Humor} experiments are sequences with a length of $4$ at most. \ac{BCE} is used as the loss function. 

\section{Baseline Results}\label{sec:results}

We implement the training of \acp{GRU} as outlined in the preceding section. In the following, we introduce and analyze the baseline results.

\subsection{\ac{MuSe-Perception}}
For the \ac{MuSe-Perception} sub-challenge, we first present the mean Pearson's correlation over all 16 target dimensions for each feature in \Cref{tab:perception}.  Averaged across all targets, \vit leads to the best overall single-modality results at $\rho=.3679$ on development and $\rho=.2577$ on test. Looking only at the audio modality, \ac{eGeMAPS} outperforms both evaluated deep features -- \ds and \wtv -- at $.2561$ and $.1442$. Overall, however, results on development and test partition often diverge substantially, \eg \ds drops from $.2075$ to $.0113$. The weighted late fusion approach, which chooses the best model for each target separately, leads to greatly increased correlations on both development ($\rho=.5579$) and test set ($\rho=.3573$), suggesting that no singular feature works best for every target label.

To further investigate how the configurations fare at recognizing each of the 16 attributes, we report the best development and accompanying test set results for the individual target attributes in~\Cref{tab:perception-single-target}. We further make use of the ``Big Two'' dimensions \emph{Agency} and \emph{Communion}~\cite{bakan_duality_1966} to split the target labels into two groups. From the \emph{agentive} subgroup, \emph{aggressive} can be recognized quite well by models trained on any of the evaluated features, with \vit achieving the best results on both development and test results and further showing solid generalization capabilities. However, for \emph{confident}, the best strong development performance of \vit does not transfer to the test set, dropping from $.7373$ to a mere $.0783$ Pearson's correlation. In the \emph{communal} subgroup, \emph{good-natured} and \emph{kind} are detected quite reliably with good generalization behavior from models trained on visual features. Audio models trained on deep features (\ds and \wtv), on the other hand, see performance drop to chance level when moving from development to test.

 \begin{table*}[ht!]
  \renewcommand{\arraystretch}{1.2}

\caption{Baseline results measured in Pearson's correlation for the \ac{MuSe-Perception} sub-challenge disaggregated by target label. For brevity, we only report the best development and accompanying test set result. Furthermore, we group the 16 aspects of social gender into ``agentive'' and ``communal'' attributes, based on~\cite{bakan_duality_1966}}\label{tab:results}\centering
\resizebox{\linewidth}{!}{
  
  \begin{tabular}{lrr|rr|rr||rr|rr|rr}
    \toprule
    & \multicolumn{12}{c}{\textbf{Evaluation Metric: }$[\rho \uparrow]$} \\
    &  \multicolumn{6}{c}{Audio} & \multicolumn{6}{c}{Visual} \\
    \cmidrule(lr){2-7} 
    \cmidrule(lr){8-13}
    Target & \multicolumn{2}{c}{\ds} & \multicolumn{2}{c}{eGeMAPS} & \multicolumn{2}{c}{\wtv} & \multicolumn{2}{c}{\facenet} & \multicolumn{2}{c}{\ac{FAU}} & \multicolumn{2}{c}{\vit}
    \\
    & Dev. & Test & Dev. & Test & Dev. & Test & Dev. & Test & Dev. & Test & Dev. & Test     
    \\ \midrule
\multicolumn{13}{l}{\textbf{\underline{agentive}}} \\ 
aggressive      & .4728  & .1323   & .3687  & .3572   & .2601  & .3266  & .2300  & .4204   & .1877  & .3123  & .4911  & .4718   \\
arrogant        & .3963  & -.1216  & .4394  & .3422   & .4503  & .4808  & .6501  & .3926   & .1717  & .3242  & .4066  & .4412   \\
assertive       & .0990  & -.0281  & .3623  & -.0334  & .3732  & .1277  & .1870  & .3528   & .4342  & -.0868 & .4139  & .2437   \\
confident       & .0730  & .3600   & .3955  & -.0594  & .4916  & .0618  & .3675  & -.0131  & .4873  & .2885  & .7373  & .0783   \\
dominant        & .1402  & .0077   & .6405  & .2408   & .5844  & .1293  & .7488  & .4434   & .5846  & .1925  & .3494  & .2977   \\
independent     & .2294  & -.0800  & .4550  & -.1849  & .5567  & .3514  & .3963  & .2789   & .4015  & .1957  & .7641  & .2398   \\
risk-taking     & .3349  & .0984   & .5166  & .4479   & .4363  & .2074  & .6402  & .3919   & .3951  & -.1371 & .5940  & .3047   \\
leader-like     & .0595  & -.2099  & .4212  & -.1064  & .6330  & .2826  & .2588  & .1135   & .2979  & .0658  & .6274  & .5135   \\
\midrule 
\multicolumn{13}{l}{\textbf{\underline{communal}}} \\ 
collaborative   & .2053  & .2049   & .3178  & .2835   & .3118  & -.1357 & .1582  & .1698   & .0668  & .2768  & .1671  & .2565  \\
enthusiastic    & .2285  & .2650   & .4505  & .2294   & .3795  & .0705  & .4545  & .0196   & .5338  & .2946  & .6151  & .1770 \\
friendly        & .3810  & .3825   & .3843  & .3052   & .3885  & .1116  & .1431  & -.1210  & .2708  & .3512  & .6428  & .3871 \\
good-natured    & .2484  & .1960   & .4424  & .1906   & .3923  & .1209  & .3340  & .3607   & .3536  & .3224  & .6596  & .5047 \\
kind            & .3987  & .0068   & .4117  & .1665   & .2808  & .0382  & .4460  & .3606   & .2515  & .3271  & .6113  & .3768 \\
likeable        & .4155  & .1034   & .3114  & .1456   & .3779  & .1423  & .2500  & -.2306  & .2605  & .2383  & .5203  & .1734 \\
sincere         & .1374  & .0984   & .4343  & .0972   & .4780  & .2541  & .1633  & -.0285  & .1967  & .3864  & .0673  & .3463 \\
warm            & .4873  & .5090   & .2266  & .2219   & .3886  & .1732  & .4337  & .4151   & .3263  & .2725  & .3927  & .2201 \\

    \bottomrule
  \end{tabular}\label{tab:perception-single-target}
  
}

\end{table*}

\begin{table*}[h!]
\renewcommand{\arraystretch}{1.2}
\caption{\ac{MuSe-Perception} baseline results. Each line refers to experiments conducted with 5 fixed seeds and reports the best Pearson correlation among them, together with the mean Pearson correlations and their standard deviations across the 5 seeds.}


\centering
 \begin{tabular}{lcc}
 \toprule 
 & \multicolumn{2}{c}{\textbf{Evaluation Metric:} $[\rho \uparrow]$} \\
 Features & \multicolumn{1}{c}{Development} & \multicolumn{1}{c}{Test}  \\ \midrule \midrule
 
 \multicolumn{3}{l}{\textbf{Audio}} \\
 \ac{eGeMAPS} & .2561 (.1931 $\pm$ .0512) & .1442 (.1424 $\pm$ .0740) \\
 \ds & .2075 (.1686 $\pm$ .0509) & .0113 (.0236 $\pm$ .0272) \\
  \wtv & .1448 (.0687 $\pm$ .0664) & .0950 (.0765 $\pm$ .0765) \\
 \midrule
 
  \multicolumn{3}{l}{\textbf{Video}} \\
 \ac{FAU} & .2143 (.1703 $\pm$ .0382) & .0793 (.0886 $\pm$ .0190) \\
 \vit & .3679 (.3042 $\pm$ .0634) & .2577 (.2333 $\pm$ .0240) \\
\facenet & .2248 (.1755 $\pm$ .0228) & .1586 (.1167 $\pm$ .0129) \\
 \midrule
 
 \multicolumn{3}{l}{\textbf{Late Fusion}} \\
 Audio + Video & .5579 (.5016 $\pm$ .0421) & \textbf{.3573} (.3122 $\pm$ .0237) \\
 \bottomrule
 \end{tabular}\label{tab:perception}

\end{table*}

\subsection{\ac{MuSe-Humor}}\label{ssec:results_humor}
\Cref{tab:humor} reports the baselines for \ac{MuSe-Humor}.

\begin{table*}[h!]
    \renewcommand{\arraystretch}{1.2}

\caption{\ac{MuSe-Humor} baseline results. Each line refers to experiments conducted with 5 fixed seeds and reports the best AUC-Score among them, together with the mean \ac{AUC}-Scores and their standard deviations across the 5 seeds.}


\centering
 \begin{tabular}{lcc}
 \toprule 
 & \multicolumn{2}{c}{\textbf{Evaluation Metric: }[\ac{AUC} $\uparrow$]} \\
 Features & \multicolumn{1}{c}{Development} & \multicolumn{1}{c}{Test}  \\ \midrule \midrule
 
 \multicolumn{3}{l}{\textbf{Audio}} \\
 \ac{eGeMAPS} & .7094 (.6717 $\pm$ .0206) & .6665 (.6610 $\pm$ .0105) \\
 \ds & .6963 (.6931 $\pm$ .0033) & .6987 (.6996 $\pm$ .0040) \\
  \wtv & .8392 (.8316 $\pm$ .0062) & .8042 (.8020 $\pm$ .0023) \\
 \midrule
 
  \multicolumn{3}{l}{\textbf{Video}} \\
 \ac{FAU} & .7749 (.7673 $\pm$ .0043) & .6400 (.5941 $\pm$ .0513) \\
 \vit & .8932 (.8888 $\pm$ .0025) & .7995 (.8052 $\pm$ .0030) \\
\facenet & .7311 (.6340 $\pm$ .0640) & .5812 (.5831 $\pm$ .0197) \\
 \midrule
 
 \multicolumn{3}{l}{\textbf{Text}} \\
 \textsc{BERT} & .8109 (.7697 $\pm$ .0680) & .7581 (.7163 $\pm$ .0886)
 \\
 \midrule
 \multicolumn{3}{l}{\textbf{Late Fusion}} \\
 Audio + Text & .8827 (.8697 $\pm$ .0194) & .8194 (.8173 $\pm$ .0114) \\
 Audio + Video & .916 (.9119 $\pm$ .0054) & .8534 (.8544 $\pm$ .0023) \\
 Text + Video & .9132 (.9059 $\pm$ .0102) & .8473 (.8384 $\pm$ .0184) \\
 Audio + Text + Video & .9251 (.9185 $\pm$ .0069) & \textbf{.8682} (.8638 $\pm$ .0079) \\
 \bottomrule
 \end{tabular}\label{tab:humor}

\end{table*}

All models achieve above-chance ($0.5$ \ac{AUC}) results, regardless of modality and employed feature representations. In this year's rendition of the sub-challenge, the introduction of \vit features helped the visual modality catch up to the best audio results found again with \wtv embeddings. \acp{GRU} trained on these features reach $.7995$ (\vit) and $.8042$ (\wtv) on the test partition. However, it can be observed that \vit features do not generalize as well to unseen data, with performance dropping from $.8932$ to $.7995$ between development and test partitions. As last year, these generalization deficits extend across all visual features while the audio modality fares better, with \wtv and \ac{eGeMAPS} only slightly dropping in \ac{AUC} and \ds matching development performance on the test set. The efficacy of exploiting linguistic information is further demonstrated by the purely textual \textsc{BERT} features, which come third after \wtv and \vit, outperforming all other single modality models.

Our late fusion approach consistently improves over the respective unimodal baselines. The performance gains achieved in the visual modality with \vit features further lead to fusion settings, which include video, eclipsing those that rely on only audio and text. The overall highest \acp{AUC} of $.9251$ $.8682$ on development and test sets, respectively, are achieved by fusing all three modalities.

\section{Conclusions}\label{sec:conclusion}
We introduced \ac{MuSe} 2024 -- the 5th Multimodal Sentiment Analysis challenge, comprising two sub-challenges, \ac{MuSe-Perception} and \ac{MuSe-Humor}.   For the \ac{MuSe-Perception} sub-challenge, the novel \ac{LMU-ELP} dataset has been introduced and is made available, which consists of interview recordings of CEOs who present their firms to potential investors before taking their firms public. Participants were tasked to predict $16$ different attributes of CEOs, \eg \emph{confidence}, or \emph{sincerity} on a Likert scale from $1$ to $7$.  

The \ac{MuSe-Humor} sub-challenge is a relaunch of the same task from the 2023 edition of \ac{MuSe}~\cite{christ2023muse,amiriparian23muse}. It uses an extended version of the \ac{Passau-SFCH} dataset~\cite{christ2022multimodal}. Participants are tasked to detect spontaneous humor in press conferences across cultures and languages, training their models on recordings of German-speaking trainers and testing on English data.

We employed publicly accessible codes and repositories to extract audio, visual, and textual features. 
Additionally, we trained simple \ac{GRU} models on the acquired representations to establish the official challenge baselines. These \textbf{baselines} represent the models' performance on the test partitions of each sub-challenge, outlined as follows: a mean $\rho$ value of \textbf{$\textbf{.3573}$ for \ac{MuSe-Perception}}, and an \ac{AUC}
value of \textbf{$\textbf{.8682}$ for \ac{MuSe-Humor}}. Both baseline results were achieved via late fusion of all modalities in each respective sub-challenge.

By sharing our code, datasets, and features publicly, we aim to facilitate broader participation and engagement from the research community. This open approach promotes transparency and encourages researchers to build upon our work, accelerating progress in multimodal data processing. As we continue to refine our baseline systems and explore new methodologies, we anticipate further advancements in our understanding of human behavior and communication across different modalities. Through ongoing collaboration and experimentation within the \ac{MuSe} 2024 framework, we hope to drive innovation and ultimately contribute to the development of more effective multimodal machine learning systems for behavioral modeling and affective analysis.

\section{Acknowledgments}
This project has received funding from the Deutsche Forschungsgemeinschaft (DFG) under grant agreement No.\ 461420398, and the DFG's Reinhart Koselleck project No.\ 442218748 (AUDI0NOMOUS). Shahin Amiriparian and Bj\"or W. Schuller are also affiliated with the Munich Center for Machine Learning (MCML), Germany. Bj\"or W. Schuller is further affiliated with the Chair of Embedded Intelligence for Healthcare and Wellbeing (EIHW), University of Augsburg, Germany, Chair of Health Informatics (CHI), Klinikum rechts der Isar (MRI), Technical University of Munich, Germany, and the Munich Data Science Institute (MDSI), Germany.

\begin{acronym}
\acro{AReLU}[AReLU]{Attention-based Rectified Linear Unit}
\acro{AUC}[AUC]{Area Under the Curve}
\acro{ASR}[ASR]{Automatic Speech Recognition}
\acro{BCE}[BCE]{Binary Cross-Entropy}
\acro{BSRI}[BSRI]{Bem Sex-Role Inventory}
\acro{CCC}[CCC]{Concordance Correlation Coefficient}
\acro{CNN}[CNN]{Convolutional Neural Network}
\acrodefplural{CNN}[CNNs]{Convolutional Neural Networks}
\acro{CI}[CI]{Confidence Interval}
\acrodefplural{CI}[CIs]{Confidence Intervals}
\acro{CCS}[CCS]{COVID-19 Cough}
\acro{CSS}[CSS]{COVID-19 Speech}
\acro{CTW}[CTW]{Canonical Time Warping}
\acro{ComParE}[ComParE]{Computational Paralinguistics Challenge}
\acrodefplural{ComParE}[ComParE]{Computational Paralinguistics Challenges}
\acro{DNN}[DNN]{Deep Neural Network}
\acrodefplural{DNNs}[DNNs]{Deep Neural Networks}
\acro{DEMoS}[DEMoS]{Database of Elicited Mood in Speech}
\acro{eGeMAPS}[\textsc{eGeMAPS}]{extended Geneva Minimalistic Acoustic Parameter Set}
\acro{EULA}[EULA]{End User License Agreement}
\acro{EWE}[EWE]{Evaluator Weighted Estimator}
\acro{FLOP}[FLOP]{Floating Point Operation}
\acrodefplural{FLOP}[FLOPs]{Floating Point Operations}
\acro{FAU}[FAU]{Facial Action Unit}
\acrodefplural{FAU}[FAUs]{Facial Action Units}
\acro{GDPR}[GDPR]{General Data Protection Regulation}
\acro{GRU}[GRU]{Gated Recurrent Unit}
\acro{HDF}[HDF]{Hierarchical Data Format}
\acro{Hume-Reaction}[\textsc{Hume-Reaction}]{Hume-Reaction}
\acro{HSQ}[HSQ]{Humor Style Questionnaire}
\acro{IEMOCAP}[IEMOCAP]{Interactive Emotional Dyadic Motion Capture}
\acro{KSS}[KSS]{Karolinska Sleepiness Scale}
\acro{LIME}[LIME]{Local Interpretable Model-agnostic Explanations}
\acro{LLD}[LLD]{Low-Level Descriptor}
\acro{LLM}[LLM]{Large Language Model}
\acro{LMU-ELP}[LMU-ELP]{LMU Munich Executive Leadership Perception}
\acrodefplural{LLD}[LLDs]{Low-Level Descriptors}
\acro{LSTM}[LSTM]{Long Short-Term Memory}
\acro{MIP}[MIP]{Mood Induction Procedure}
\acro{MIP}[MIPs]{Mood Induction Procedures}
\acro{MLP}[MLP]{Multilayer Perceptron}
\acrodefplural{MLP}[MLPs]{Multilayer Perceptrons}
\acro{MPSSC}[MPSSC]{Munich-Passau Snore Sound Corpus}
\acro{MSE}[MSE]{Mean Squared Error}
\acro{MTCNN}[MTCNN]{Multi-task Cascaded Convolutional Networks}
\acro{MuSe}[MuSe]{\textbf{Mu}ltimodal \textbf{Se}ntiment Analysis Challenge}
\acro{MuSe-Humor}[\textsc{MuSe-Humor}]{Cross-Cultural Humor Detection Sub-Challenge}
\acro{MuSe-Mimic}[\textsc{MuSe-Mimic}]{Mimicked Emotions Sub-Challenge}
\acro{MuSe-Stress}[\textsc{MuSe-Stress}]{Emotional Stress Sub-Challenge}
\acro{MuSe-Personalisation}[\textsc{MuSe-Personalisation}]{Personalisation Sub-Challenge}
\acro{MuSe-Perception}[\textsc{MuSe-Perception}]{Social Perception Sub-Challenge}
\acro{Passau-SFCH}[\textsc{Passau-SFCH}]{Passau Spontaneous Football Coach Humor}
\acro{RAAW}[\textsc{RAAW}]{Rater Aligned Annotation Weighting}
\acro{RAVDESS}[RAVDESS]{Ryerson Audio-Visual Database of Emotional Speech and Song}
\acro{RNN}[RNN]{Recurrent Neural Network}
\acro{SER}[SER]{Speech Emotion Recognition}
\acro{SHAP}[SHAP]{SHapley Additive exPlanations}
\acro{SLEEP}[SLEEP]{Düsseldorf Sleepy Language Corpus}
\acro{STFT}[STFT]{Short-Time Fourier Transform}
\acrodefplural{STFT}[STFTs]{Short-Time Fourier Transforms}
\acro{SVM}[SVM]{Support Vector Machine}
\acro{TF}[TF]{TensorFlow}
\acro{TSST}[TSST]{Trier Social Stress Test}
\acro{TNR}[TNR]{True Negative Rate}
\acro{TPR}[TPR]{True Positive Rate}
\acro{UAR}[UAR]{Unweighted Average Recall}
\acro{Ulm-TSST}[\textsc{Ulm-TSST}]{Ulm-Trier Social Stress Test}
\acrodefplural{UAR}[UARs]{Unweighted Average Recall}
\end{acronym}

\clearpage
\footnotesize
\bibliographystyle{ACM-Reference-Format}
\balance
\bibliography{sample-base}


\begin{thebibliography}{72}


\ifx \showCODEN    \undefined \def \showCODEN     #1{\unskip}     \fi
\ifx \showDOI      \undefined \def \showDOI       #1{#1}\fi
\ifx \showISBNx    \undefined \def \showISBNx     #1{\unskip}     \fi
\ifx \showISBNxiii \undefined \def \showISBNxiii  #1{\unskip}     \fi
\ifx \showISSN     \undefined \def \showISSN      #1{\unskip}     \fi
\ifx \showLCCN     \undefined \def \showLCCN      #1{\unskip}     \fi
\ifx \shownote     \undefined \def \shownote      #1{#1}          \fi
\ifx \showarticletitle \undefined \def \showarticletitle #1{#1}   \fi
\ifx \showURL      \undefined \def \showURL       {\relax}        \fi
\providecommand\bibfield[2]{#2}
\providecommand\bibinfo[2]{#2}
\providecommand\natexlab[1]{#1}
\providecommand\showeprint[2][]{arXiv:#2}

\bibitem[Abele et~al\mbox{.}(2021)]%
        {abele2021navigating}
\bibfield{author}{\bibinfo{person}{Andrea~E Abele}, \bibinfo{person}{Naomi Ellemers}, \bibinfo{person}{Susan~T Fiske}, \bibinfo{person}{Alex Koch}, {and} \bibinfo{person}{Vincent Yzerbyt}.} \bibinfo{year}{2021}\natexlab{}.
\newblock \showarticletitle{Navigating the social world: Toward an integrated framework for evaluating self, individuals, and groups.}
\newblock \bibinfo{journal}{\emph{Psychological Review}} \bibinfo{volume}{128}, \bibinfo{number}{2} (\bibinfo{year}{2021}), \bibinfo{pages}{290}.
\newblock


\bibitem[Abele and Wojciszke(2014)]%
        {abele_chapter_2014}
\bibfield{author}{\bibinfo{person}{Andrea~E. Abele} {and} \bibinfo{person}{Bogdan Wojciszke}.} \bibinfo{year}{2014}\natexlab{}.
\newblock \showarticletitle{Chapter {Four} - {Communal} and {Agentic} {Content} in {Social} {Cognition}: {A} {Dual} {Perspective} {Model}}.
\newblock In \bibinfo{booktitle}{\emph{Advances in {Experimental} {Social} {Psychology}}}, \bibfield{editor}{\bibinfo{person}{James~M. Olson} {and} \bibinfo{person}{Mark~P. Zanna}} (Eds.). Vol.~\bibinfo{volume}{50}. \bibinfo{publisher}{Academic Press}, \bibinfo{pages}{195--255}.
\newblock
\urldef\tempurl%
\url{https://doi.org/10.1016/B978-0-12-800284-1.00004-7}
\showDOI{\tempurl}


\bibitem[Ambady and Skowronski(2008)]%
        {ambady2008first}
\bibfield{author}{\bibinfo{person}{Nalini Ambady} {and} \bibinfo{person}{John~Joseph Skowronski}.} \bibinfo{year}{2008}\natexlab{}.
\newblock \bibinfo{booktitle}{\emph{First impressions}}.
\newblock \bibinfo{publisher}{Guilford Press}.
\newblock


\bibitem[Amiriparian et~al\mbox{.}(2023)]%
        {amiriparian23muse}
\bibfield{author}{\bibinfo{person}{Shahin Amiriparian}, \bibinfo{person}{Lukas Christ}, \bibinfo{person}{Andreas K\"onig}, \bibinfo{person}{Eva-Maria Me{ss}ner}, \bibinfo{person}{Alan Cowen}, \bibinfo{person}{Erik Cambria}, {and} \bibinfo{person}{Bj\"orn~W. Schuller}.} \bibinfo{year}{2023}\natexlab{}.
\newblock \showarticletitle{MuSe 2023 Challenge: Multimodal Prediction of Mimicked Emotions, Cross-Cultural Humour, and Personalised Recognition of Affects}. In \bibinfo{booktitle}{\emph{Proceedings of the 31st ACM International Conference on Multimedia (MM'23), October 29-November 2, 2023, Ottawa, Canada.}} \bibinfo{publisher}{Association for Computing Machinery}, \bibinfo{address}{Ottawa, Canada}.
\newblock
\newblock
\shownote{to appear}.


\bibitem[Amiriparian et~al\mbox{.}(2017a)]%
        {Amiriparian17-SAU}
\bibfield{author}{\bibinfo{person}{Shahin Amiriparian}, \bibinfo{person}{Nicholas Cummins}, \bibinfo{person}{Sandra Ottl}, \bibinfo{person}{Maurice Gerczuk}, {and} \bibinfo{person}{Bj\"orn Schuller}.} \bibinfo{year}{2017}\natexlab{a}.
\newblock \showarticletitle{{Sentiment Analysis Using Image-based Deep Spectrum Features}}. In \bibinfo{booktitle}{\emph{{Proceedings 2nd International Workshop on Automatic Sentiment Analysis in the Wild (WASA 2017) held in conjunction with the 7th biannual Conference on Affective Computing and Intelligent Interaction (ACII 2017)}}}. AAAC, \bibinfo{publisher}{IEEE}, \bibinfo{address}{San Antonio, TX}, \bibinfo{pages}{26--29}.
\newblock


\bibitem[Amiriparian et~al\mbox{.}(2017b)]%
        {Amiriparian17-SSC}
\bibfield{author}{\bibinfo{person}{Shahin Amiriparian}, \bibinfo{person}{Maurice Gerczuk}, \bibinfo{person}{Sandra Ottl}, \bibinfo{person}{Nicholas Cummins}, \bibinfo{person}{Michael Freitag}, \bibinfo{person}{Sergey Pugachevskiy}, {and} \bibinfo{person}{Bj\"orn Schuller}.} \bibinfo{year}{2017}\natexlab{b}.
\newblock \showarticletitle{{Snore Sound Classification Using Image-based Deep Spectrum Features}}. In \bibinfo{booktitle}{\emph{{Proceedings INTERSPEECH 2017, 18th Annual Conference of the International Speech Communication Association}}}. ISCA, \bibinfo{publisher}{ISCA}, \bibinfo{address}{Stockholm, Sweden}, \bibinfo{pages}{3512--3516}.
\newblock


\bibitem[Amiriparian et~al\mbox{.}(2020)]%
        {Amiriparian20-TCP}
\bibfield{author}{\bibinfo{person}{Shahin Amiriparian}, \bibinfo{person}{Maurice Gerczuk}, \bibinfo{person}{Lukas Stappen}, \bibinfo{person}{Alice Baird}, \bibinfo{person}{Lukas Koebe}, \bibinfo{person}{Sandra Ottl}, {and} \bibinfo{person}{Bj\"orn Schuller}.} \bibinfo{year}{2020}\natexlab{}.
\newblock \showarticletitle{{Towards Cross-Modal Pre-Training and Learning Tempo-Spatial Characteristics for Audio Recognition with Convolutional and Recurrent Neural Networks}}.
\newblock \bibinfo{journal}{\emph{EURASIP Journal on Audio, Speech, and Music Processing}} \bibinfo{volume}{2020}, \bibinfo{number}{19} (\bibinfo{year}{2020}), \bibinfo{pages}{1--11}.
\newblock


\bibitem[Amiriparian et~al\mbox{.}(2022)]%
        {Amiriparian22-DAP}
\bibfield{author}{\bibinfo{person}{Shahin Amiriparian}, \bibinfo{person}{Tobias Hübner}, \bibinfo{person}{Vincent Karas}, \bibinfo{person}{Maurice Gerczuk}, \bibinfo{person}{Sandra Ottl}, {and} \bibinfo{person}{Björn~W. Schuller}.} \bibinfo{year}{2022}\natexlab{}.
\newblock \showarticletitle{DeepSpectrumLite: A Power-Efficient Transfer Learning Framework for Embedded Speech and Audio Processing From Decentralized Data}.
\newblock \bibinfo{journal}{\emph{Frontiers in Artificial Intelligence}}  \bibinfo{volume}{5} (\bibinfo{year}{2022}), \bibinfo{numpages}{10}~pages.
\newblock
\showISSN{2624-8212}
\urldef\tempurl%
\url{https://doi.org/10.3389/frai.2022.856232}
\showDOI{\tempurl}


\bibitem[Baevski et~al\mbox{.}(2020)]%
        {baevski2020wav2vec}
\bibfield{author}{\bibinfo{person}{Alexei Baevski}, \bibinfo{person}{Yuhao Zhou}, \bibinfo{person}{Abdelrahman Mohamed}, {and} \bibinfo{person}{Michael Auli}.} \bibinfo{year}{2020}\natexlab{}.
\newblock \showarticletitle{wav2vec 2.0: A framework for self-supervised learning of speech representations}.
\newblock \bibinfo{journal}{\emph{Advances in neural information processing systems}}  \bibinfo{volume}{33} (\bibinfo{year}{2020}), \bibinfo{pages}{12449--12460}.
\newblock


\bibitem[Baird et~al\mbox{.}(2019)]%
        {baird2019can}
\bibfield{author}{\bibinfo{person}{Alice Baird}, \bibinfo{person}{Shahin Amiriparian}, {and} \bibinfo{person}{Bj{\"o}rn Schuller}.} \bibinfo{year}{2019}\natexlab{}.
\newblock \showarticletitle{Can deep generative audio be emotional? Towards an approach for personalised emotional audio generation}. In \bibinfo{booktitle}{\emph{2019 IEEE 21st International Workshop on Multimedia Signal Processing (MMSP)}}. IEEE, \bibinfo{publisher}{IEEE}, \bibinfo{address}{Kuala Lumpur, Malaysia}, \bibinfo{pages}{1--5}.
\newblock


\bibitem[Bakan(1966)]%
        {bakan_duality_1966}
\bibfield{author}{\bibinfo{person}{David Bakan}.} \bibinfo{year}{1966}\natexlab{}.
\newblock \bibinfo{booktitle}{\emph{The duality of human existence: {An} essay on psychology and religion}}.
\newblock \bibinfo{publisher}{Rand Mcnally}, \bibinfo{address}{Oxford, England}.
\newblock
\newblock
\shownote{Pages: 242}.


\bibitem[Bem(1981)]%
        {bem1981manual}
\bibfield{author}{\bibinfo{person}{Sandra~L Bem}.} \bibinfo{year}{1981}\natexlab{}.
\newblock \showarticletitle{A manual for the Bem sex role inventory}.
\newblock \bibinfo{journal}{\emph{California: Mind Garden}} (\bibinfo{year}{1981}).
\newblock


\bibitem[Bertero and Fung(2016)]%
        {bertero2016deep}
\bibfield{author}{\bibinfo{person}{Dario Bertero} {and} \bibinfo{person}{Pascale Fung}.} \bibinfo{year}{2016}\natexlab{}.
\newblock \showarticletitle{Deep learning of audio and language features for humor prediction}. In \bibinfo{booktitle}{\emph{Proceedings of the Tenth International Conference on Language Resources and Evaluation (LREC'16)}}. \bibinfo{pages}{496--501}.
\newblock


\bibitem[{Bj\"orn W.\ Schuller and Anton Batliner and Christian Bergler and Cecilia Mascolo and Jing Han and Iulia Lefter and Heysem Kaya and Shahin Amiriparian and Alice Baird and Lukas Stappen and Sandra Ottl and Maurice Gerczuk and Panaguiotis Tzirakis and Chlo\"e Brown and Jagmohan Chauhan and Andreas Grammenos and Apinan Hasthanasombat and Dimitris Spathis and Tong Xia and Pietro Cicuta and Leon J.\,M.\ Rothkrantz and Joeri Zwerts and Jelle Treep and Casper Kaandorp}(2021)]%
        {Schuller21-TI2}
\bibfield{author}{\bibinfo{person}{{Bj\"orn W.\ Schuller and Anton Batliner and Christian Bergler and Cecilia Mascolo and Jing Han and Iulia Lefter and Heysem Kaya and Shahin Amiriparian and Alice Baird and Lukas Stappen and Sandra Ottl and Maurice Gerczuk and Panaguiotis Tzirakis and Chlo\"e Brown and Jagmohan Chauhan and Andreas Grammenos and Apinan Hasthanasombat and Dimitris Spathis and Tong Xia and Pietro Cicuta and Leon J.\,M.\ Rothkrantz and Joeri Zwerts and Jelle Treep and Casper Kaandorp}}.} \bibinfo{year}{2021}\natexlab{}.
\newblock \showarticletitle{{The INTERSPEECH 2021 Computational Paralinguistics Challenge: COVID-19 Cough, COVID-19 Speech, Escalation \& Primates}}. In \bibinfo{booktitle}{\emph{{Proceedings INTERSPEECH 2021, 22nd Annual Conference of the International Speech Communication Association}}}. ISCA, \bibinfo{publisher}{ISCA}, \bibinfo{address}{Brno, Czechia}, \bibinfo{pages}{431--435}.
\newblock


\bibitem[Breil et~al\mbox{.}(2021)]%
        {breil202113}
\bibfield{author}{\bibinfo{person}{Simon~M Breil}, \bibinfo{person}{Sarah Osterholz}, \bibinfo{person}{Steffen Nestler}, {and} \bibinfo{person}{Mitja~D Back}.} \bibinfo{year}{2021}\natexlab{}.
\newblock \showarticletitle{13 contributions of nonverbal cues to the accurate judgment of personality traits}.
\newblock \bibinfo{journal}{\emph{The Oxford handbook of accurate personality judgment}} (\bibinfo{year}{2021}), \bibinfo{pages}{195--218}.
\newblock


\bibitem[Calder et~al\mbox{.}(2011)]%
        {calder2011personality}
\bibfield{author}{\bibinfo{person}{Andrew~J Calder}, \bibinfo{person}{Michael Ewbank}, {and} \bibinfo{person}{Luca Passamonti}.} \bibinfo{year}{2011}\natexlab{}.
\newblock \showarticletitle{Personality influences the neural responses to viewing facial expressions of emotion}.
\newblock \bibinfo{journal}{\emph{Philosophical Transactions of the Royal Society B: Biological Sciences}} \bibinfo{volume}{366}, \bibinfo{number}{1571} (\bibinfo{year}{2011}), \bibinfo{pages}{1684--1701}.
\newblock


\bibitem[Cann et~al\mbox{.}(2014)]%
        {cann2014assessing}
\bibfield{author}{\bibinfo{person}{Arnie Cann}, \bibinfo{person}{Amanda~J Watson}, {and} \bibinfo{person}{Elisabeth~A Bridgewater}.} \bibinfo{year}{2014}\natexlab{}.
\newblock \showarticletitle{Assessing humor at work: The humor climate questionnaire}.
\newblock \bibinfo{journal}{\emph{Humor}} \bibinfo{volume}{27}, \bibinfo{number}{2} (\bibinfo{year}{2014}), \bibinfo{pages}{307--323}.
\newblock


\bibitem[Chaudhari et~al\mbox{.}(2022)]%
        {chaudhari2022vitfer}
\bibfield{author}{\bibinfo{person}{Aayushi Chaudhari}, \bibinfo{person}{Chintan Bhatt}, \bibinfo{person}{Achyut Krishna}, {and} \bibinfo{person}{Pier~Luigi Mazzeo}.} \bibinfo{year}{2022}\natexlab{}.
\newblock \showarticletitle{ViTFER: facial emotion recognition with vision transformers}.
\newblock \bibinfo{journal}{\emph{Applied System Innovation}} \bibinfo{volume}{5}, \bibinfo{number}{4} (\bibinfo{year}{2022}), \bibinfo{pages}{80}.
\newblock


\bibitem[Chen and Zhang(2022)]%
        {chen2022integrating}
\bibfield{author}{\bibinfo{person}{Chengxin Chen} {and} \bibinfo{person}{Pengyuan Zhang}.} \bibinfo{year}{2022}\natexlab{}.
\newblock \showarticletitle{Integrating Cross-Modal Interactions via Latent Representation Shift for Multi-Modal Humor Detection}. In \bibinfo{booktitle}{\emph{Proceedings of the 3rd International on Multimodal Sentiment Analysis Workshop and Challenge}} (Lisboa, Portugal) \emph{(\bibinfo{series}{MuSe' 22})}. \bibinfo{publisher}{Association for Computing Machinery}, \bibinfo{address}{New York, NY, USA}, \bibinfo{pages}{23–28}.
\newblock
\showISBNx{9781450394840}
\urldef\tempurl%
\url{https://doi.org/10.1145/3551876.3554805}
\showDOI{\tempurl}


\bibitem[Christ et~al\mbox{.}(2023a)]%
        {christ2023muse}
\bibfield{author}{\bibinfo{person}{Lukas Christ}, \bibinfo{person}{Shahin Amiriparian}, \bibinfo{person}{Alice Baird}, \bibinfo{person}{Alexander Kathan}, \bibinfo{person}{Niklas M{\"u}ller}, \bibinfo{person}{Steffen Klug}, \bibinfo{person}{Chris Gagne}, \bibinfo{person}{Panagiotis Tzirakis}, \bibinfo{person}{Lukas Stappen}, \bibinfo{person}{Eva-Maria Me{\ss}ner}, {et~al\mbox{.}}} \bibinfo{year}{2023}\natexlab{a}.
\newblock \showarticletitle{The muse 2023 multimodal sentiment analysis challenge: Mimicked emotions, cross-cultural humour, and personalisation}. In \bibinfo{booktitle}{\emph{Proceedings of the 4th on Multimodal Sentiment Analysis Challenge and Workshop: Mimicked Emotions, Humour and Personalisation}}. \bibinfo{pages}{1--10}.
\newblock


\bibitem[Christ et~al\mbox{.}(2023b)]%
        {christ2022multimodal}
\bibfield{author}{\bibinfo{person}{Lukas Christ}, \bibinfo{person}{Shahin Amiriparian}, \bibinfo{person}{Alexander Kathan}, \bibinfo{person}{Niklas M{\"u}ller}, \bibinfo{person}{Andreas K{\"o}nig}, {and} \bibinfo{person}{Bj{\"o}rn~W Schuller}.} \bibinfo{year}{2023}\natexlab{b}.
\newblock \showarticletitle{Towards Multimodal Prediction of Spontaneous Humour: A Novel Dataset and First Results}.
\newblock \bibinfo{journal}{\emph{arXiv preprint arXiv:2209.14272}} (\bibinfo{year}{2023}).
\newblock


\bibitem[Cutler and Condon(2022)]%
        {cutler2022deep}
\bibfield{author}{\bibinfo{person}{Andrew Cutler} {and} \bibinfo{person}{David~M Condon}.} \bibinfo{year}{2022}\natexlab{}.
\newblock \showarticletitle{Deep lexical hypothesis: Identifying personality structure in natural language.}
\newblock \bibinfo{journal}{\emph{Journal of Personality and Social Psychology}} (\bibinfo{year}{2022}).
\newblock


\bibitem[Devlin et~al\mbox{.}(2019)]%
        {devlin-etal-2019-bert}
\bibfield{author}{\bibinfo{person}{Jacob Devlin}, \bibinfo{person}{Ming-Wei Chang}, \bibinfo{person}{Kenton Lee}, {and} \bibinfo{person}{Kristina Toutanova}.} \bibinfo{year}{2019}\natexlab{}.
\newblock \showarticletitle{{BERT}: Pre-training of Deep Bidirectional Transformers for Language Understanding}. In \bibinfo{booktitle}{\emph{Proceedings of the 2019 Conference of the North {A}merican Chapter of the Association for Computational Linguistics: Human Language Technologies}}. \bibinfo{pages}{4171--4186}.
\newblock


\bibitem[Dosovitskiy et~al\mbox{.}(2021)]%
        {dosovitskiy2021an}
\bibfield{author}{\bibinfo{person}{Alexey Dosovitskiy}, \bibinfo{person}{Lucas Beyer}, \bibinfo{person}{Alexander Kolesnikov}, \bibinfo{person}{Dirk Weissenborn}, \bibinfo{person}{Xiaohua Zhai}, \bibinfo{person}{Thomas Unterthiner}, \bibinfo{person}{Mostafa Dehghani}, \bibinfo{person}{Matthias Minderer}, \bibinfo{person}{Georg Heigold}, \bibinfo{person}{Sylvain Gelly}, \bibinfo{person}{Jakob Uszkoreit}, {and} \bibinfo{person}{Neil Houlsby}.} \bibinfo{year}{2021}\natexlab{}.
\newblock \showarticletitle{An Image is Worth 16x16 Words: Transformers for Image Recognition at Scale}. In \bibinfo{booktitle}{\emph{International Conference on Learning Representations}}.
\newblock
\urldef\tempurl%
\url{https://openreview.net/forum?id=YicbFdNTTy}
\showURL{%
\tempurl}


\bibitem[Eagly and Karau(2002)]%
        {eagly2002role}
\bibfield{author}{\bibinfo{person}{Alice~H Eagly} {and} \bibinfo{person}{Steven~J Karau}.} \bibinfo{year}{2002}\natexlab{}.
\newblock \showarticletitle{Role congruity theory of prejudice toward female leaders.}
\newblock \bibinfo{journal}{\emph{Psychological review}} \bibinfo{volume}{109}, \bibinfo{number}{3} (\bibinfo{year}{2002}), \bibinfo{pages}{573}.
\newblock


\bibitem[Ekman and Friesen(1978)]%
        {ekman1978facial}
\bibfield{author}{\bibinfo{person}{Paul Ekman} {and} \bibinfo{person}{Wallace~V Friesen}.} \bibinfo{year}{1978}\natexlab{}.
\newblock \showarticletitle{Facial action coding system}.
\newblock \bibinfo{journal}{\emph{Environmental Psychology \& Nonverbal Behavior}} (\bibinfo{year}{1978}).
\newblock


\bibitem[Eulitz and Gazdag(2021)]%
        {eulitz_beyond_2021}
\bibfield{author}{\bibinfo{person}{Simone~Maria Eulitz} {and} \bibinfo{person}{Brooke~A. Gazdag}.} \bibinfo{year}{2021}\natexlab{}.
\newblock \showarticletitle{Beyond {Biology} – {The} impact of {Perceptions} {CEO} {Social} {Gender} on {Investor} {Reactions} {During} an {IPO}}.
\newblock \bibinfo{journal}{\emph{Academy of Management Proceedings}} \bibinfo{volume}{2021}, \bibinfo{number}{1} (\bibinfo{date}{Aug.} \bibinfo{year}{2021}), \bibinfo{pages}{12379}.
\newblock
\showISSN{0065-0668}
\urldef\tempurl%
\url{https://doi.org/10.5465/AMBPP.2021.12379abstract}
\showDOI{\tempurl}
\newblock
\shownote{Publisher: Academy of Management}.


\bibitem[Eyben et~al\mbox{.}(2015)]%
        {eyben2015geneva}
\bibfield{author}{\bibinfo{person}{Florian Eyben}, \bibinfo{person}{Klaus~R Scherer}, \bibinfo{person}{Bj{\"o}rn~W Schuller}, \bibinfo{person}{Johan Sundberg}, \bibinfo{person}{Elisabeth Andr{\'e}}, \bibinfo{person}{Carlos Busso}, \bibinfo{person}{Laurence~Y Devillers}, \bibinfo{person}{Julien Epps}, \bibinfo{person}{Petri Laukka}, \bibinfo{person}{Shrikanth~S Narayanan}, {et~al\mbox{.}}} \bibinfo{year}{2015}\natexlab{}.
\newblock \showarticletitle{The Geneva minimalistic acoustic parameter set (GeMAPS) for voice research and affective computing}.
\newblock \bibinfo{journal}{\emph{IEEE Transactions on Affective Computing}} \bibinfo{volume}{7}, \bibinfo{number}{2} (\bibinfo{year}{2015}), \bibinfo{pages}{190--202}.
\newblock


\bibitem[Eyben et~al\mbox{.}(2010)]%
        {eyben2010opensmile}
\bibfield{author}{\bibinfo{person}{Florian Eyben}, \bibinfo{person}{Martin W{\"o}llmer}, {and} \bibinfo{person}{Bj{\"o}rn Schuller}.} \bibinfo{year}{2010}\natexlab{}.
\newblock \showarticletitle{Opensmile: the munich versatile and fast open-source audio feature extractor}. In \bibinfo{booktitle}{\emph{Proceedings of the 18th ACM International Conference on Multimedia}}. \bibinfo{publisher}{Association for Computing Machinery}, \bibinfo{address}{Firenze, Italy}, \bibinfo{pages}{1459--1462}.
\newblock


\bibitem[Gerczuk et~al\mbox{.}(2022)]%
        {Gerczuk22-EAT}
\bibfield{author}{\bibinfo{person}{Maurice Gerczuk}, \bibinfo{person}{Shahin Amiriparian}, \bibinfo{person}{Sandra Ottl}, {and} \bibinfo{person}{Bj\"orn Schuller}.} \bibinfo{year}{2022}\natexlab{}.
\newblock \showarticletitle{{EmoNet: A Transfer Learning Framework for Multi-Corpus Speech Emotion Recognition}}.
\newblock \bibinfo{journal}{\emph{IEEE Transactions on Affective Computing}}  \bibinfo{volume}{13} (\bibinfo{year}{2022}).
\newblock


\bibitem[Gkorezis et~al\mbox{.}(2014)]%
        {gkorezis2014leader}
\bibfield{author}{\bibinfo{person}{Panagiotis Gkorezis}, \bibinfo{person}{Eugenia Petridou}, {and} \bibinfo{person}{Panteleimon Xanthiakos}.} \bibinfo{year}{2014}\natexlab{}.
\newblock \showarticletitle{Leader positive humor and organizational cynicism: LMX as a mediator}.
\newblock \bibinfo{journal}{\emph{Leadership \& Organization Development Journal}}  \bibinfo{volume}{35} (\bibinfo{year}{2014}), \bibinfo{pages}{305 -- 315}.
\newblock


\bibitem[Goodfellow et~al\mbox{.}(2013)]%
        {goodfellow2013challenges}
\bibfield{author}{\bibinfo{person}{Ian~J Goodfellow}, \bibinfo{person}{Dumitru Erhan}, \bibinfo{person}{Pierre~Luc Carrier}, \bibinfo{person}{Aaron Courville}, \bibinfo{person}{Mehdi Mirza}, \bibinfo{person}{Ben Hamner}, \bibinfo{person}{Will Cukierski}, \bibinfo{person}{Yichuan Tang}, \bibinfo{person}{David Thaler}, \bibinfo{person}{Dong-Hyun Lee}, {et~al\mbox{.}}} \bibinfo{year}{2013}\natexlab{}.
\newblock \showarticletitle{Challenges in representation learning: A report on three machine learning contests}. In \bibinfo{booktitle}{\emph{Neural Information Processing: 20th International Conference, ICONIP 2013, Daegu, Korea, November 3-7, 2013. Proceedings, Part III 20}}. Springer, \bibinfo{pages}{117--124}.
\newblock


\bibitem[Gr{\'o}sz et~al\mbox{.}(2023)]%
        {grosz2023discovering}
\bibfield{author}{\bibinfo{person}{Tam{\'a}s Gr{\'o}sz}, \bibinfo{person}{Anja Virkkunen}, \bibinfo{person}{Dejan Porjazovski}, {and} \bibinfo{person}{Mikko Kurimo}.} \bibinfo{year}{2023}\natexlab{}.
\newblock \showarticletitle{Discovering Relevant Sub-spaces of BERT, Wav2Vec 2.0, ELECTRA and ViT Embeddings for Humor and Mimicked Emotion Recognition with Integrated Gradients}. In \bibinfo{booktitle}{\emph{Proceedings of the 4th on Multimodal Sentiment Analysis Challenge and Workshop: Mimicked Emotions, Humour and Personalisation}}. \bibinfo{pages}{27--34}.
\newblock


\bibitem[Guhr et~al\mbox{.}(2021)]%
        {guhr-EtAl:2021:fullstop}
\bibfield{author}{\bibinfo{person}{Oliver Guhr}, \bibinfo{person}{Anne-Kathrin Schumann}, \bibinfo{person}{Frank Bahrmann}, {and} \bibinfo{person}{Hans~Joachim Böhme}.} \bibinfo{year}{2021}\natexlab{}.
\newblock \showarticletitle{FullStop: Multilingual Deep Models for Punctuation Prediction}. In \bibinfo{booktitle}{\emph{Proceedings of the Swiss Text Analytics Conference 2021}}. \bibinfo{publisher}{CEUR Workshop Proceedings}, \bibinfo{address}{Winterthur, Switzerland}.
\newblock
\urldef\tempurl%
\url{http://ceur-ws.org/Vol-2957/sepp_paper4.pdf}
\showURL{%
\tempurl}


\bibitem[Hasan et~al\mbox{.}(2021)]%
        {hasan2021humor}
\bibfield{author}{\bibinfo{person}{Md~Kamrul Hasan}, \bibinfo{person}{Sangwu Lee}, \bibinfo{person}{Wasifur Rahman}, \bibinfo{person}{Amir Zadeh}, \bibinfo{person}{Rada Mihalcea}, \bibinfo{person}{Louis-Philippe Morency}, {and} \bibinfo{person}{Ehsan Hoque}.} \bibinfo{year}{2021}\natexlab{}.
\newblock \showarticletitle{Humor knowledge enriched transformer for understanding multimodal humor}. In \bibinfo{booktitle}{\emph{Proceedings of the AAAI Conference on Artificial Intelligence}}, Vol.~\bibinfo{volume}{35}. \bibinfo{pages}{12972--12980}.
\newblock


\bibitem[Hasan et~al\mbox{.}(2019)]%
        {hasan2019ur}
\bibfield{author}{\bibinfo{person}{Md~Kamrul Hasan}, \bibinfo{person}{Wasifur Rahman}, \bibinfo{person}{AmirAli Bagher~Zadeh}, \bibinfo{person}{Jianyuan Zhong}, \bibinfo{person}{Md~Iftekhar Tanveer}, \bibinfo{person}{Louis-Philippe Morency}, {and} \bibinfo{person}{Mohammed~(Ehsan) Hoque}.} \bibinfo{year}{2019}\natexlab{}.
\newblock \showarticletitle{{UR}-{FUNNY}: A Multimodal Language Dataset for Understanding Humor}. In \bibinfo{booktitle}{\emph{Proceedings of the 2019 Conference on Empirical Methods in Natural Language Processing and the 9th International Joint Conference on Natural Language Processing (EMNLP-IJCNLP)}}. \bibinfo{publisher}{Association for Computational Linguistics}, \bibinfo{address}{Hong Kong, China}, \bibinfo{pages}{2046--2056}.
\newblock
\urldef\tempurl%
\url{https://doi.org/10.18653/v1/D19-1211}
\showDOI{\tempurl}


\bibitem[Jiang et~al\mbox{.}(2023)]%
        {jiang2023mistral}
\bibfield{author}{\bibinfo{person}{Albert~Q Jiang}, \bibinfo{person}{Alexandre Sablayrolles}, \bibinfo{person}{Arthur Mensch}, \bibinfo{person}{Chris Bamford}, \bibinfo{person}{Devendra~Singh Chaplot}, \bibinfo{person}{Diego de~las Casas}, \bibinfo{person}{Florian Bressand}, \bibinfo{person}{Gianna Lengyel}, \bibinfo{person}{Guillaume Lample}, \bibinfo{person}{Lucile Saulnier}, {et~al\mbox{.}}} \bibinfo{year}{2023}\natexlab{}.
\newblock \showarticletitle{Mistral 7B}.
\newblock \bibinfo{journal}{\emph{arXiv preprint arXiv:2310.06825}} (\bibinfo{year}{2023}).
\newblock


\bibitem[Kachur et~al\mbox{.}(2020)]%
        {kachur2020assessing}
\bibfield{author}{\bibinfo{person}{Alexander Kachur}, \bibinfo{person}{Evgeny Osin}, \bibinfo{person}{Denis Davydov}, \bibinfo{person}{Konstantin Shutilov}, {and} \bibinfo{person}{Alexey Novokshonov}.} \bibinfo{year}{2020}\natexlab{}.
\newblock \showarticletitle{Assessing the Big Five personality traits using real-life static facial images}.
\newblock \bibinfo{journal}{\emph{Scientific Reports}} \bibinfo{volume}{10}, \bibinfo{number}{1} (\bibinfo{year}{2020}), \bibinfo{pages}{8487}.
\newblock


\bibitem[Ladilova and Schr{\"o}der(2022)]%
        {ladilova2022humor}
\bibfield{author}{\bibinfo{person}{Anna Ladilova} {and} \bibinfo{person}{Ulrike Schr{\"o}der}.} \bibinfo{year}{2022}\natexlab{}.
\newblock \showarticletitle{Humor in intercultural interaction: A source for misunderstanding or a common ground builder? A multimodal analysis}.
\newblock \bibinfo{journal}{\emph{Intercultural Pragmatics}} \bibinfo{volume}{19}, \bibinfo{number}{1} (\bibinfo{year}{2022}), \bibinfo{pages}{71--101}.
\newblock


\bibitem[Li et~al\mbox{.}(2022)]%
        {li2022hybrid}
\bibfield{author}{\bibinfo{person}{Jia Li}, \bibinfo{person}{Ziyang Zhang}, \bibinfo{person}{Junjie Lang}, \bibinfo{person}{Yueqi Jiang}, \bibinfo{person}{Liuwei An}, \bibinfo{person}{Peng Zou}, \bibinfo{person}{Yangyang Xu}, \bibinfo{person}{Sheng Gao}, \bibinfo{person}{Jie Lin}, \bibinfo{person}{Chunxiao Fan}, \bibinfo{person}{Xiao Sun}, {and} \bibinfo{person}{Meng Wang}.} \bibinfo{year}{2022}\natexlab{}.
\newblock \showarticletitle{Hybrid Multimodal Feature Extraction, Mining and Fusion for Sentiment Analysis}. In \bibinfo{booktitle}{\emph{Proceedings of the 3rd International on Multimodal Sentiment Analysis Workshop and Challenge}} (Lisboa, Portugal) \emph{(\bibinfo{series}{MuSe' 22})}. \bibinfo{publisher}{Association for Computing Machinery}, \bibinfo{address}{New York, NY, USA}, \bibinfo{pages}{81–88}.
\newblock
\showISBNx{9781450394840}
\urldef\tempurl%
\url{https://doi.org/10.1145/3551876.3554809}
\showDOI{\tempurl}


\bibitem[Li et~al\mbox{.}(2023a)]%
        {li2023temporal}
\bibfield{author}{\bibinfo{person}{Qi Li}, \bibinfo{person}{Shulei Tang}, \bibinfo{person}{Feixiang Zhang}, \bibinfo{person}{Ruotong Wang}, \bibinfo{person}{Yangyang Xu}, \bibinfo{person}{Zhuoer Zhao}, \bibinfo{person}{Xiao Sun}, {and} \bibinfo{person}{Meng Wang}.} \bibinfo{year}{2023}\natexlab{a}.
\newblock \showarticletitle{Temporal-aware Multimodal Feature Fusion for Sentiment Analysis}. In \bibinfo{booktitle}{\emph{Proceedings of the 4th on Multimodal Sentiment Analysis Challenge and Workshop: Mimicked Emotions, Humour and Personalisation}}. \bibinfo{pages}{99--105}.
\newblock


\bibitem[Li et~al\mbox{.}(2023b)]%
        {li2023jtma}
\bibfield{author}{\bibinfo{person}{Qi Li}, \bibinfo{person}{Yangyang Xu}, \bibinfo{person}{Zhuoer Zhao}, \bibinfo{person}{Shulei Tang}, \bibinfo{person}{Feixiang Zhang}, \bibinfo{person}{Ruotong Wang}, \bibinfo{person}{Xiao Sun}, {and} \bibinfo{person}{Meng Wang}.} \bibinfo{year}{2023}\natexlab{b}.
\newblock \showarticletitle{JTMA: Joint Multimodal Feature Fusion and Temporal Multi-head Attention for Humor Detection}. In \bibinfo{booktitle}{\emph{Proceedings of the 4th on Multimodal Sentiment Analysis Challenge and Workshop: Mimicked Emotions, Humour and Personalisation}}. \bibinfo{pages}{59--65}.
\newblock


\bibitem[Liu et~al\mbox{.}(2022)]%
        {ssl-survey}
\bibfield{author}{\bibinfo{person}{Shuo Liu}, \bibinfo{person}{Adria Mallol-Ragolta}, \bibinfo{person}{Emilia Parada-Cabaleiro}, \bibinfo{person}{Kun Qian}, \bibinfo{person}{Xin Jing}, \bibinfo{person}{Alexander Kathan}, \bibinfo{person}{Bin Hu}, {and} \bibinfo{person}{Björn~W. Schuller}.} \bibinfo{year}{2022}\natexlab{}.
\newblock \showarticletitle{Audio self-supervised learning: A survey}.
\newblock \bibinfo{journal}{\emph{Patterns}} \bibinfo{volume}{3}, \bibinfo{number}{12} (\bibinfo{year}{2022}), \bibinfo{pages}{100616}.
\newblock
\showISSN{2666-3899}
\urldef\tempurl%
\url{https://doi.org/10.1016/j.patter.2022.100616}
\showDOI{\tempurl}


\bibitem[Martin et~al\mbox{.}(2003)]%
        {martin2003individual}
\bibfield{author}{\bibinfo{person}{Rod~A Martin}, \bibinfo{person}{Patricia Puhlik-Doris}, \bibinfo{person}{Gwen Larsen}, \bibinfo{person}{Jeanette Gray}, {and} \bibinfo{person}{Kelly Weir}.} \bibinfo{year}{2003}\natexlab{}.
\newblock \showarticletitle{Individual differences in uses of humor and their relation to psychological well-being: Development of the Humor Styles Questionnaire}.
\newblock \bibinfo{journal}{\emph{Journal of research in personality}} \bibinfo{volume}{37}, \bibinfo{number}{1} (\bibinfo{year}{2003}), \bibinfo{pages}{48--75}.
\newblock


\bibitem[McAuliffe et~al\mbox{.}(2017)]%
        {mcauliffe2017montreal}
\bibfield{author}{\bibinfo{person}{Michael McAuliffe}, \bibinfo{person}{Michaela Socolof}, \bibinfo{person}{Sarah Mihuc}, \bibinfo{person}{Michael Wagner}, {and} \bibinfo{person}{Morgan Sonderegger}.} \bibinfo{year}{2017}\natexlab{}.
\newblock \showarticletitle{Montreal Forced Aligner: Trainable Text-Speech Alignment Using Kaldi.}. In \bibinfo{booktitle}{\emph{Proceedings of INTERSPEECH}}, Vol.~\bibinfo{volume}{2017}. \bibinfo{publisher}{International Speech Communication Association (ISCA)}, \bibinfo{address}{Stockholm, Sweden}, \bibinfo{pages}{498--502}.
\newblock


\bibitem[Mittal et~al\mbox{.}(2021)]%
        {mittal2021so}
\bibfield{author}{\bibinfo{person}{Anirudh Mittal}, \bibinfo{person}{Pranav Jeevan}, \bibinfo{person}{Prerak Gandhi}, \bibinfo{person}{Diptesh Kanojia}, {and} \bibinfo{person}{Pushpak Bhattacharyya}.} \bibinfo{year}{2021}\natexlab{}.
\newblock \bibinfo{title}{" So You Think You're Funny?": Rating the Humour Quotient in Standup Comedy}.
\newblock
\newblock
\showeprint{arXiv preprint arXiv:2110.12765}


\bibitem[Morais et~al\mbox{.}(2022)]%
        {morais2022speech}
\bibfield{author}{\bibinfo{person}{Edmilson Morais}, \bibinfo{person}{Ron Hoory}, \bibinfo{person}{Weizhong Zhu}, \bibinfo{person}{Itai Gat}, \bibinfo{person}{Matheus Damasceno}, {and} \bibinfo{person}{Hagai Aronowitz}.} \bibinfo{year}{2022}\natexlab{}.
\newblock \showarticletitle{Speech emotion recognition using self-supervised features}. In \bibinfo{booktitle}{\emph{ICASSP 2022-2022 IEEE International Conference on Acoustics, Speech and Signal Processing (ICASSP)}}. IEEE, \bibinfo{pages}{6922--6926}.
\newblock


\bibitem[Ottl et~al\mbox{.}(2020)]%
        {Ottl20-GSE}
\bibfield{author}{\bibinfo{person}{Sandra Ottl}, \bibinfo{person}{Shahin Amiriparian}, \bibinfo{person}{Maurice Gerczuk}, \bibinfo{person}{Vincent Karas}, {and} \bibinfo{person}{Bj\"orn Schuller}.} \bibinfo{year}{2020}\natexlab{}.
\newblock \showarticletitle{{Group-level Speech Emotion Recognition Utilising Deep Spectrum Features}}. In \bibinfo{booktitle}{\emph{{Proceedings of the 8th ICMI 2020 EmotiW -- Emotion Recognition In The Wild Challenge (EmotiW 2020), 22nd ACM International Conference on Multimodal Interaction (ICMI 2020)}}}. ACM, \bibinfo{publisher}{ACM}, \bibinfo{address}{Utrecht, The Netherlands}, \bibinfo{pages}{821--826}.
\newblock


\bibitem[Park et~al\mbox{.}(2023)]%
        {park2023muse}
\bibfield{author}{\bibinfo{person}{Ho-Min Park}, \bibinfo{person}{Ganghyun Kim}, \bibinfo{person}{Arnout Van~Messem}, {and} \bibinfo{person}{Wesley De~Neve}.} \bibinfo{year}{2023}\natexlab{}.
\newblock \showarticletitle{MuSe-Personalization 2023: Feature Engineering, Hyperparameter Optimization, and Transformer-Encoder Re-discovery}. In \bibinfo{booktitle}{\emph{Proceedings of the 4th on Multimodal Sentiment Analysis Challenge and Workshop: Mimicked Emotions, Humour and Personalisation}}. \bibinfo{pages}{89--97}.
\newblock


\bibitem[Park et~al\mbox{.}(2022)]%
        {park2022towards}
\bibfield{author}{\bibinfo{person}{Ho-min Park}, \bibinfo{person}{Ilho Yun}, \bibinfo{person}{Ajit Kumar}, \bibinfo{person}{Ankit~Kumar Singh}, \bibinfo{person}{Bong~Jun Choi}, \bibinfo{person}{Dhananjay Singh}, {and} \bibinfo{person}{Wesley De~Neve}.} \bibinfo{year}{2022}\natexlab{}.
\newblock \showarticletitle{Towards Multimodal Prediction of Time-continuous Emotion using Pose Feature Engineering and a Transformer Encoder}. In \bibinfo{booktitle}{\emph{Proceedings of the 3rd International on Multimodal Sentiment Analysis Workshop and Challenge}}. \bibinfo{pages}{47--54}.
\newblock


\bibitem[Pepino et~al\mbox{.}(2021)]%
        {pepino21_interspeech}
\bibfield{author}{\bibinfo{person}{Leonardo Pepino}, \bibinfo{person}{Pablo Riera}, {and} \bibinfo{person}{Luciana Ferrer}.} \bibinfo{year}{2021}\natexlab{}.
\newblock \showarticletitle{{Emotion Recognition from Speech Using wav2vec 2.0 Embeddings}}. In \bibinfo{booktitle}{\emph{Proc. Interspeech 2021}}. ISCA, \bibinfo{publisher}{ISCA}, \bibinfo{address}{Brno, Czechia}, \bibinfo{pages}{3400--3404}.
\newblock
\urldef\tempurl%
\url{https://doi.org/10.21437/Interspeech.2021-703}
\showDOI{\tempurl}


\bibitem[Pires et~al\mbox{.}(2019)]%
        {pires-etal-2019-multilingual}
\bibfield{author}{\bibinfo{person}{Telmo Pires}, \bibinfo{person}{Eva Schlinger}, {and} \bibinfo{person}{Dan Garrette}.} \bibinfo{year}{2019}\natexlab{}.
\newblock \showarticletitle{How Multilingual is Multilingual {BERT}?}. In \bibinfo{booktitle}{\emph{Proceedings of the 57th Annual Meeting of the Association for Computational Linguistics}}. \bibinfo{publisher}{Association for Computational Linguistics}, \bibinfo{address}{Florence, Italy}, \bibinfo{pages}{4996--5001}.
\newblock
\urldef\tempurl%
\url{https://doi.org/10.18653/v1/P19-1493}
\showDOI{\tempurl}


\bibitem[Pradhan et~al\mbox{.}(2020)]%
        {pradhan2020analysis}
\bibfield{author}{\bibinfo{person}{Tejas Pradhan}, \bibinfo{person}{Rashi Bhansali}, \bibinfo{person}{Dimple Chandnani}, {and} \bibinfo{person}{Aditya Pangaonkar}.} \bibinfo{year}{2020}\natexlab{}.
\newblock \showarticletitle{Analysis of personality traits using natural language processing and deep learning}. In \bibinfo{booktitle}{\emph{2020 Second International Conference on Inventive Research in Computing Applications (ICIRCA)}}. IEEE, \bibinfo{pages}{457--461}.
\newblock


\bibitem[Pramanick et~al\mbox{.}(2022)]%
        {pramanick2022multimodal}
\bibfield{author}{\bibinfo{person}{Shraman Pramanick}, \bibinfo{person}{Aniket Roy}, {and} \bibinfo{person}{Vishal~M Patel}.} \bibinfo{year}{2022}\natexlab{}.
\newblock \showarticletitle{Multimodal Learning using Optimal Transport for Sarcasm and Humor Detection}. In \bibinfo{booktitle}{\emph{Proceedings of the IEEE/CVF Winter Conference on Applications of Computer Vision}}. \bibinfo{pages}{3930--3940}.
\newblock


\bibitem[Priego-Valverde et~al\mbox{.}(2018)]%
        {priego2018smiling}
\bibfield{author}{\bibinfo{person}{B{\'e}atrice Priego-Valverde}, \bibinfo{person}{Brigitte Bigi}, \bibinfo{person}{Salvatore Attardo}, \bibinfo{person}{Lucy Pickering}, {and} \bibinfo{person}{Elisa Gironzetti}.} \bibinfo{year}{2018}\natexlab{}.
\newblock \showarticletitle{Is smiling during humor so obvious? a cross-cultural comparison of smiling behavior in humorous sequences in american english and french interactions}.
\newblock \bibinfo{journal}{\emph{Intercultural Pragmatics}} \bibinfo{volume}{15}, \bibinfo{number}{4} (\bibinfo{year}{2018}), \bibinfo{pages}{563--591}.
\newblock


\bibitem[Sadvilkar and Neumann(2020)]%
        {sadvilkar-neumann-2020-pysbd}
\bibfield{author}{\bibinfo{person}{Nipun Sadvilkar} {and} \bibinfo{person}{Mark Neumann}.} \bibinfo{year}{2020}\natexlab{}.
\newblock \showarticletitle{{P}y{SBD}: Pragmatic Sentence Boundary Disambiguation}. In \bibinfo{booktitle}{\emph{Proceedings of Second Workshop for NLP Open Source Software (NLP-OSS)}}. \bibinfo{publisher}{Association for Computational Linguistics}, \bibinfo{address}{Online}, \bibinfo{pages}{110--114}.
\newblock
\urldef\tempurl%
\url{https://www.aclweb.org/anthology/2020.nlposs-1.15}
\showURL{%
\tempurl}


\bibitem[Schroff et~al\mbox{.}(2015)]%
        {schroff2015facenet}
\bibfield{author}{\bibinfo{person}{Florian Schroff}, \bibinfo{person}{Dmitry Kalenichenko}, {and} \bibinfo{person}{James Philbin}.} \bibinfo{year}{2015}\natexlab{}.
\newblock \showarticletitle{FaceNet: A unified embedding for face recognition and clustering}. In \bibinfo{booktitle}{\emph{2015 IEEE Conference on Computer Vision and Pattern Recognition (CVPR)}}. \bibinfo{publisher}{IEEE}.
\newblock
\urldef\tempurl%
\url{https://doi.org/10.1109/cvpr.2015.7298682}
\showDOI{\tempurl}


\bibitem[Serengil and Ozpinar(2020)]%
        {serengil2020lightface}
\bibfield{author}{\bibinfo{person}{Sefik~Ilkin Serengil} {and} \bibinfo{person}{Alper Ozpinar}.} \bibinfo{year}{2020}\natexlab{}.
\newblock \showarticletitle{LightFace: A Hybrid Deep Face Recognition Framework}. In \bibinfo{booktitle}{\emph{2020 Innovations in Intelligent Systems and Applications Conference (ASYU)}}. IEEE, \bibinfo{pages}{23--27}.
\newblock
\urldef\tempurl%
\url{https://doi.org/10.1109/ASYU50717.2020.9259802}
\showDOI{\tempurl}


\bibitem[Sonlu et~al\mbox{.}(2021)]%
        {sonlu2021conversational}
\bibfield{author}{\bibinfo{person}{Sinan Sonlu}, \bibinfo{person}{U{\u{g}}ur G{\"u}d{\"u}kbay}, {and} \bibinfo{person}{Funda Durupinar}.} \bibinfo{year}{2021}\natexlab{}.
\newblock \showarticletitle{A conversational agent framework with multi-modal personality expression}.
\newblock \bibinfo{journal}{\emph{ACM Transactions on Graphics (TOG)}} \bibinfo{volume}{40}, \bibinfo{number}{1} (\bibinfo{year}{2021}), \bibinfo{pages}{1--16}.
\newblock


\bibitem[Taylor and Mazlack(2004)]%
        {taylor2004computationally}
\bibfield{author}{\bibinfo{person}{Julia~M Taylor} {and} \bibinfo{person}{Lawrence~J Mazlack}.} \bibinfo{year}{2004}\natexlab{}.
\newblock \showarticletitle{Computationally recognizing wordplay in jokes}. In \bibinfo{booktitle}{\emph{Proceedings of the Annual Meeting of the Cognitive Science Society}}, Vol.~\bibinfo{volume}{26}.
\newblock


\bibitem[Touvron et~al\mbox{.}(2023)]%
        {touvron2023llama}
\bibfield{author}{\bibinfo{person}{Hugo Touvron}, \bibinfo{person}{Thibaut Lavril}, \bibinfo{person}{Gautier Izacard}, \bibinfo{person}{Xavier Martinet}, \bibinfo{person}{Marie-Anne Lachaux}, \bibinfo{person}{Timoth{\'e}e Lacroix}, \bibinfo{person}{Baptiste Rozi{\`e}re}, \bibinfo{person}{Naman Goyal}, \bibinfo{person}{Eric Hambro}, \bibinfo{person}{Faisal Azhar}, {et~al\mbox{.}}} \bibinfo{year}{2023}\natexlab{}.
\newblock \showarticletitle{Llama: Open and efficient foundation language models}.
\newblock \bibinfo{journal}{\emph{arXiv preprint arXiv:2302.13971}} (\bibinfo{year}{2023}).
\newblock


\bibitem[Vlasenko et~al\mbox{.}(2021)]%
        {vlasenko2021fusion}
\bibfield{author}{\bibinfo{person}{Bogdan Vlasenko}, \bibinfo{person}{RaviShankar Prasad}, {and} \bibinfo{person}{Mathew Magimai.-Doss}.} \bibinfo{year}{2021}\natexlab{}.
\newblock \showarticletitle{Fusion of Acoustic and Linguistic Information using Supervised Autoencoder for Improved Emotion Recognition}.
\newblock In \bibinfo{booktitle}{\emph{Proceedings of the 2nd on Multimodal Sentiment Analysis Challenge}}. \bibinfo{publisher}{Association for Computing Machinery}, \bibinfo{address}{New York, NY, USA}, \bibinfo{pages}{51--59}.
\newblock


\bibitem[Wagner et~al\mbox{.}(2023)]%
        {wagner2022dawn}
\bibfield{author}{\bibinfo{person}{J. Wagner}, \bibinfo{person}{A. Triantafyllopoulos}, \bibinfo{person}{H. Wierstorf}, \bibinfo{person}{M. Schmitt}, \bibinfo{person}{F. Burkhardt}, \bibinfo{person}{F. Eyben}, {and} \bibinfo{person}{B.~W. Schuller}.} \bibinfo{year}{2023}\natexlab{}.
\newblock \showarticletitle{Dawn of the Transformer Era in Speech Emotion Recognition: Closing the Valence Gap}.
\newblock \bibinfo{journal}{\emph{IEEE Transactions on Pattern Analysis \& Machine Intelligence}} \bibinfo{number}{01} (\bibinfo{year}{2023}), \bibinfo{pages}{1--13}.
\newblock
\showISSN{1939-3539}
\urldef\tempurl%
\url{https://doi.org/10.1109/TPAMI.2023.3263585}
\showDOI{\tempurl}


\bibitem[Wu et~al\mbox{.}(2021)]%
        {wu2021mumor}
\bibfield{author}{\bibinfo{person}{Jiaming Wu}, \bibinfo{person}{Hongfei Lin}, \bibinfo{person}{Liang Yang}, {and} \bibinfo{person}{Bo Xu}.} \bibinfo{year}{2021}\natexlab{}.
\newblock \showarticletitle{MUMOR: A Multimodal Dataset for Humor Detection in Conversations}. In \bibinfo{booktitle}{\emph{CCF International Conference on Natural Language Processing and Chinese Computing}}. Springer, \bibinfo{publisher}{Springer}, \bibinfo{address}{Qingdao, China}, \bibinfo{pages}{619--627}.
\newblock


\bibitem[Xie et~al\mbox{.}(2023)]%
        {xie2023multimodal}
\bibfield{author}{\bibinfo{person}{Heng Xie}, \bibinfo{person}{Jizhou Cui}, \bibinfo{person}{Yuhang Cao}, \bibinfo{person}{Junjie Chen}, \bibinfo{person}{Jianhua Tao}, \bibinfo{person}{Cunhang Fan}, \bibinfo{person}{Xuefei Liu}, \bibinfo{person}{Zhengqi Wen}, \bibinfo{person}{Heng Lu}, \bibinfo{person}{Yuguang Yang}, {et~al\mbox{.}}} \bibinfo{year}{2023}\natexlab{}.
\newblock \showarticletitle{Multimodal Cross-Lingual Features and Weight Fusion for Cross-Cultural Humor Detection}. In \bibinfo{booktitle}{\emph{Proceedings of the 4th on Multimodal Sentiment Analysis Challenge and Workshop: Mimicked Emotions, Humour and Personalisation}}. \bibinfo{pages}{51--57}.
\newblock


\bibitem[Xu et~al\mbox{.}(2022)]%
        {xu2022hybrid}
\bibfield{author}{\bibinfo{person}{Haojie Xu}, \bibinfo{person}{Weifeng Liu}, \bibinfo{person}{Jiangwei Liu}, \bibinfo{person}{Mingzheng Li}, \bibinfo{person}{Yu Feng}, \bibinfo{person}{Yasi Peng}, \bibinfo{person}{Yunwei Shi}, \bibinfo{person}{Xiao Sun}, {and} \bibinfo{person}{Meng Wang}.} \bibinfo{year}{2022}\natexlab{}.
\newblock \showarticletitle{Hybrid Multimodal Fusion for Humor Detection}. In \bibinfo{booktitle}{\emph{Proceedings of the 3rd International on Multimodal Sentiment Analysis Workshop and Challenge}} (Lisboa, Portugal) \emph{(\bibinfo{series}{MuSe' 22})}. \bibinfo{publisher}{Association for Computing Machinery}, \bibinfo{address}{New York, NY, USA}, \bibinfo{pages}{15–21}.
\newblock
\urldef\tempurl%
\url{https://doi.org/10.1145/3551876.3554802}
\showDOI{\tempurl}


\bibitem[Xu et~al\mbox{.}(2023)]%
        {xu2023humor}
\bibfield{author}{\bibinfo{person}{Mingyu Xu}, \bibinfo{person}{Shun Chen}, \bibinfo{person}{Zheng Lian}, {and} \bibinfo{person}{Bin Liu}.} \bibinfo{year}{2023}\natexlab{}.
\newblock \showarticletitle{Humor Detection System for MuSE 2023: Contextual Modeling, Pesudo Labelling, and Post-smoothing}. In \bibinfo{booktitle}{\emph{Proceedings of the 4th on Multimodal Sentiment Analysis Challenge and Workshop: Mimicked Emotions, Humour and Personalisation}}. \bibinfo{pages}{35--41}.
\newblock


\bibitem[Yang et~al\mbox{.}(2015)]%
        {yang2015humor}
\bibfield{author}{\bibinfo{person}{Diyi Yang}, \bibinfo{person}{Alon Lavie}, \bibinfo{person}{Chris Dyer}, {and} \bibinfo{person}{Eduard Hovy}.} \bibinfo{year}{2015}\natexlab{}.
\newblock \showarticletitle{Humor recognition and humor anchor extraction}. In \bibinfo{booktitle}{\emph{Proceedings of the 2015 conference on empirical methods in natural language processing}}. \bibinfo{publisher}{Association for Computational Linguistics}, \bibinfo{address}{Lisbon, Portugal}, \bibinfo{pages}{2367--2376}.
\newblock


\bibitem[Yi et~al\mbox{.}(2023)]%
        {yi2023exploring}
\bibfield{author}{\bibinfo{person}{Guofeng Yi}, \bibinfo{person}{Yuguang Yang}, \bibinfo{person}{Yu Pan}, \bibinfo{person}{Yuhang Cao}, \bibinfo{person}{Jixun Yao}, \bibinfo{person}{Xiang Lv}, \bibinfo{person}{Cunhang Fan}, \bibinfo{person}{Zhao Lv}, \bibinfo{person}{Jianhua Tao}, \bibinfo{person}{Shan Liang}, {et~al\mbox{.}}} \bibinfo{year}{2023}\natexlab{}.
\newblock \showarticletitle{Exploring the Power of Cross-Contextual Large Language Model in Mimic Emotion Prediction}. In \bibinfo{booktitle}{\emph{Proceedings of the 4th on Multimodal Sentiment Analysis Challenge and Workshop: Mimicked Emotions, Humour and Personalisation}}. \bibinfo{pages}{19--26}.
\newblock


\bibitem[Yu et~al\mbox{.}(2023)]%
        {yu2023mmt}
\bibfield{author}{\bibinfo{person}{Jun Yu}, \bibinfo{person}{Wangyuan Zhu}, \bibinfo{person}{Jichao Zhu}, \bibinfo{person}{Xiaxin Shen}, \bibinfo{person}{Jianqing Sun}, {and} \bibinfo{person}{Jiaen Liang}.} \bibinfo{year}{2023}\natexlab{}.
\newblock \showarticletitle{MMT-GD: Multi-Modal Transformer with Graph Distillation for Cross-Cultural Humor Detection}. In \bibinfo{booktitle}{\emph{Proceedings of the 4th on Multimodal Sentiment Analysis Challenge and Workshop: Mimicked Emotions, Humour and Personalisation}}. \bibinfo{pages}{43--49}.
\newblock


\bibitem[Zhang et~al\mbox{.}(2016)]%
        {zhang2016mtcnn}
\bibfield{author}{\bibinfo{person}{Kaipeng Zhang}, \bibinfo{person}{Zhanpeng Zhang}, \bibinfo{person}{Zhifeng Li}, {and} \bibinfo{person}{Yu Qiao}.} \bibinfo{year}{2016}\natexlab{}.
\newblock \showarticletitle{Joint Face Detection and Alignment Using Multitask Cascaded Convolutional Networks}.
\newblock \bibinfo{journal}{\emph{IEEE Signal Processing Letters}}  \bibinfo{volume}{23} (\bibinfo{date}{04} \bibinfo{year}{2016}).
\newblock


\bibitem[Zhi et~al\mbox{.}(2020)]%
        {zhi2020comprehensive}
\bibfield{author}{\bibinfo{person}{Ruicong Zhi}, \bibinfo{person}{Mengyi Liu}, {and} \bibinfo{person}{Dezheng Zhang}.} \bibinfo{year}{2020}\natexlab{}.
\newblock \showarticletitle{A comprehensive survey on automatic facial action unit analysis}.
\newblock \bibinfo{journal}{\emph{The Visual Computer}}  \bibinfo{volume}{36} (\bibinfo{year}{2020}), \bibinfo{pages}{1067--1093}.
\newblock


\end{thebibliography}

\end{document}